# PDDL2.1 : An Extension to PDDL for Expressing Temporal Planning Domains

**Maria Fox**                                               MARIA.FOX@CIS.STRATH.AC.UK
**Derek Long**                                              DEREK.LONG@CIS.STRATH.AC.UK
*Department of Computer and Information Sciences*
*University of Strathclyde, Glasgow, UK*

## Abstract

In recent years research in the planning community has moved increasingly towards application of planners to realistic problems involving both time and many types of resources. For example, interest in planning demonstrated by the space research community has inspired work in observation scheduling, planetary rover exploration and spacecraft control domains. Other temporal and resource-intensive domains including logistics planning, plant control and manufacturing have also helped to focus the community on the modelling and reasoning issues that must be confronted to make planning technology meet the challenges of application.

The International Planning Competitions have acted as an important motivating force behind the progress that has been made in planning since 1998. The third competition (held in 2002) set the planning community the challenge of handling time and numeric resources. This necessitated the development of a modelling language capable of expressing temporal and numeric properties of planning domains. In this paper we describe the language, PDDL2.1, that was used in the competition. We describe the syntax of the language, its formal semantics and the validation of concurrent plans. We observe that PDDL2.1 has considerable modelling power — exceeding the capabilities of current planning technology — and presents a number of important challenges to the research community.

## 1. Introduction

In 1998 Drew McDermott released a Planning Domain Description Language, PDDL (McDermott, 2000; McDermott & the AIPS-98 Planning Competition Committee, 1998), which has since become a community standard for the representation and exchange of planning domain models. Despite some dissatisfaction in the community with some of the features of PDDL the language has enabled considerable progress to be made in planning research because of the ease with which systems sharing the standard can be compared and the enormous increase in availability of shared planning resources. The introduction of PDDL has facilitated the scientific development of planning.

Since 1998 there has been a decisive movement in the research community towards application of planning technology to realistic problems. The propositional puzzle domains of old are no longer considered adequate for demonstrating the utility of a planning system — modern planners must be able to reason about time and numeric quantities. Although several members of the community have been working on applications of planning to real domains of this nature for some time (Laborie & Ghallab, 1995; Ghallab & Laruelle, 1994; Muscettola, 1994; Drabble & Tate, 1994; Wilkins, 1988) there has always been a gap between the modelling requirements of such domains and what can be expressed in PDDL. Application-driven planners come equipped with their own modelling conventions and black arts and, as a consequence, it is difficult to reproduce their results and to make empirical comparisons with other approaches, both of which are essential for scientific progress to be made.

The PDDL language provides the foundation on which an expressive standard can be constructed, enabling the domain models of the applications-driven community to be shared and motivating the development of the planning field towards realistic application. The third International Planning Competition, which took place in 2002, had the objective of closing the gap between planning research and application. As organisers of the third competition the authors therefore took the first step in defining an expressive language capable of modelling a certain class of temporal and resource-intensive planning domains. This had to be done both with an eye to the future and with awareness of the current capabilities of planners (it had to be possible for the language to be used by members of the community, or there would be no competitors). In this paper we describe the resulting language, PDDL2.1, in terms of its syntax, semantics and modelling capabilities.

PDDL2.1 has been designed to be backward compatible with the fragment of PDDL that has been in common usage since 1998. This compatibility supports the development of resources which help to establish a scientific foundation for the field of AI planning. Furthermore, McDermott's original PDDL provides a clean and well-understood basis for development and embodies a number of design principles that we considered it important to retain. PDDL2.1 extends PDDL in principled ways to achieve the additional expressive power following, as far as possible, McDermott's maxim "physics, not advice" (McDermott, 2000). We take this maxim to mean that a language should focus on expressing the physical properties of the world, not advice to the planner about how to search the associated solution spaces. Of course, any model of physical systems makes simplifying assumptions and abstracts behaviours at some level, so no model can be claimed to be purely physics and free of decisions that could influence the use of the model. We do not attempt to make strong judgements about what constitutes advice but try to implement the maxim by keeping the language as simple as possible. We make the following two guarantees of backward compatibility:

1. All existing PDDL domains (in common usage) are valid PDDL2.1 domains. This is important to enable existing libraries of benchmark problems to remain valid.

2. Valid PDDL plans are valid PDDL2.1 plans.

An important contribution made in the development of PDDL2.1 is a means by which domain designers can provide alternative objective functions that can be used to judge the value of a plan. The use of numbers in a domain provides a platform for measuring consumption of critical resources and other parameters. An example of a metric that can be modelled is that energy consumption must be minimized. This is very important for many practical applications of planning in which plan quality might be dependent on a number of interacting domain-dependent factors.

The organisation of the paper is as follows. In Section 2 we introduce non-specialist readers to the PDDL domain description language used in the planning research community.

This background is given in order to provide the foundations for the numeric and durative extensions made in developing PDDL2.1. The paper then focusses on the specific extensions introduced: numeric expressions and durative actions. In Section 3 we start by explaining the syntax of numeric expressions and their use in action descriptions. We then explain, in Section 4, how *metrics* can be provided as part of the problem description so that the quality of a plan involving numeric change can be evaluated in terms appropriate to the problem domain. We present the syntax in which metrics are expressed and give examples.

In Section 5 the paper introduces the notion of *durative* action as a way of modelling the temporal properties of a planning domain. Both discretised and continuous durative actions are considered. The syntax is described and examples of modelling power and limitations are presented in both cases. Having given examples of the syntactic representation of durative actions we present a formal semantics for both discretised and continuous actions and for plans. Sections 6, 7, 8 and 9 provide the details. The semantics gives us a way of tackling the problem of confirming plan validity — something that becomes an important issue in the face of concurrent activity. In Section 10 we describe the process by which plans were validated in the competition and discuss the complexity of the validation question for PDDL2.1. Finally, Section 11 describes some related work in the temporal reasoning community, in order to put the contributions made by PDDL2.1 into a wider context. A full BNF description of PDDL2.1 can be found in the appendix.

PDDL2.1 was developed for use in the third International Planning Competition in which competing planners demonstrated that many discretised temporal and metric models can now be efficiently handled by both domain-independent planners and those using hand-tailored control rules. For ease of reference in the competition we identified the features of PDDL2.1 with a series of *levels* of increasing expressive power. Thus, the STRIPS fragment of PDDL2.1 was referred to as level 1, the numeric extensions comprised level 2, the addition of discretised durative actions resulted in level 3, continuous durative actions resulted in level 4 and a final level, level 5, comprised all of the extensions of PDDL2.1 and additional components to support the modelling of spontaneous events and physical processes. Level 5 is not discussed in this paper but details can be found in earlier work by Fox and Long (2002). The competition focussed on the use of levels 1, 2 and 3 and did not use levels 4 or 5 because the planning technology was not at that stage sufficiently advanced to handle the additional complexities. Despite the fact that level 4 was not used in the competition we devote some discussion to it in this paper. We feel that level 4 presents some important immediate challenges for the planning community that affect the extent to which planning can be applied to real problems.

The purpose of this paper is to provide an overview of the new features introduced in PDDL2.1, discuss the rationale for our language choices and explain some of the issues that have arisen in trying to extend PDDL. Although we have provided the BNF for PDDL2.1 as an appendix, this paper is not intended to be either a language manual or a tutorial on the use of the language. For examples of the use of the language and other relevant materials, readers should consult archived resources currently held at `http://www.dur.ac.uk/d.p.long/competition.html`.

## 2. PDDL Background

PDDL is an action-centred language, inspired by the well-known STRIPS formulations of planning problems. At its core is a simple standardisation of the syntax for expressing this familiar semantics of actions, using pre- and post-conditions to describe the applicability and effects of actions. The syntax is inspired by Lisp, so much of the structure of a domain description is a Lisp-like list of parenthesised expressions. An early design decision in the language was to separate the descriptions of parameterised actions that characterise domain behaviours from the description of specific objects, initial conditions and goals that characterise a problem instance. Thus, a planning problem is created by the pairing of a domain description with a problem description. The same domain description can be paired with many different problem descriptions to yield different planning problems in the same domain. The parameterisation of actions depends on the use of variables that stand for terms of the problem instance — they are instantiated to objects from a specific problem instance when an action is grounded for application. The pre- and post-conditions of actions are expressed as logical propositions constructed from predicates and argument terms (objects from a problem instance) and logical connectives.

Although the core of PDDL is a STRIPS formalism, the language extends beyond that. The extended expressive power includes the ability to express a type structure for the objects in a domain, typing the parameters that appear in actions and constraining the types of arguments to predicates, actions with negative preconditions and conditional effects and the use of quantification in expressing both pre- and post-conditions. These extensions are essentially those proposed as ADL (Pednault, 1989).

Although the original definition of the PDDL syntax was not accompanied by a formal semantics, the language was really a proposal for a standard syntax for a commonly accepted semantics and there was little scope for disagreement about the meaning of the language constructs. Two parts of the original language proposal for which this claim fails are an attempt to offer a standard syntax for describing hierarchical domain descriptions, suitable for HTN planners and the subset of the language concerned with expressing numeric-valued fluents. The former was an ambitious project to construct a syntax in which the entire structure of domains using hierarchical action decompositions could be expressed. In contrast to STRIPS-based planning, the differences between planners using hierarchical decomposition appear to be deeper, with domain descriptions often containing structures that go beyond the description of domain behaviours (for example, SHOP (Nau, Cao, Lotem, & Muñoz-Avila, 1999) often uses mechanisms that represent goal agendas and other solution-oriented structures in a domain encoding). This diversity undermined the efforts at standardisation in hierarchical domain descriptions and this part of the language has never been successfully explored.

The syntax proposed for expressing numeric-valued fluents was not tested in the first use of the language (in the 1998 competition) and, indeed, it underwent revision in the early development of the language. The second competition in 2000 also avoided use of numeric-valued fluents, so a general agreement about the syntax and semantics of the numeric-expressivity of the language remained unnecessary. McDermott's original PDDL provides support for numbers by allowing numeric quantities to be assigned and updated. The syntax of numeric-valued fluents changed between the PDDL manuals 1.1 and 1.2 (McDermott &

```
(define (domain jug-pouring)
  (:requirements :typing :fluents)
  (:types jug)
  (:functors
        (amount ?j -jug)
        (capacity ?j -jug)
          - (fluent number))
  (:action empty
        :parameters (?jug1 ?jug2 - jug)
        :precondition (fluent-test
               (>= (- (capacity ?jug2) (amount ?jug2))
                      (amount ?jug1)))
        :effect (and (change (amount ?jug1) 0)
                    (change (amount ?jug2)
                        (+ (amount ?jug1) (amount ?jug2)))))))
)
```

Figure 1: Pouring water between jugs as described in the AI Magazine article (McDermott, 2000).

the AIPS-98 Planning Competition Committee, 1998) and the later AI Magazine article on PDDL (McDermott, 2000). McDermott presented a version of numeric fluents used in PDDL in the article in AI Magazine (2000) which could be taken as a definitive statement of the syntax. An example using numeric fluents, presented by McDermott (2000), is shown in Figure 1. This action models an action from the well-known jugs-and-water problem, allowing the water in one jug to be emptied into a second jug provided that the space in the second jug is large enough to hold the water in the first. The effect is a discrete update of the values of the current contents of the jugs by an assignment (denoted here by the change token).

Even without the numeric extensions, PDDL is an expressive language, capable of capturing a wide variety of interesting and challenging behaviours. Figure 2 illustrates how PDDL can be used to capture a domain in which a vehicle can move between locations, consuming fuel as it does so.

It can be seen in the example that PDDL includes a syntactic representation of the level of expressivity required in particular domain descriptions through the use of requirements flags. This gives the opportunity for a planning system to gracefully reject attempts to plan with domains that make use of more advanced features of the language than the planner can handle. Syntax checking tools can be used to confirm that the requirements flags are correctly set for a domain and that the types and other features of the language are correctly employed. An example of a problem description to accompany the vehicle domain is shown in Figure 3. The example illustrates that the description of an initial state requires an exhaustive listing of all the (atomic) propositions that hold. Symmetric or transitive relations must be modelled by exhaustive and explicit listing of the propositions that hold. The use of domain axioms to simplify the description of domains that use such relationships has been considered, but remains an untested part of PDDL and therefore

```
(define (domain vehicle)
 (:requirements :strips :typing)
 (:types vehicle location fuel-level)
 (:predicates (at ?v - vehicle ?p - location)
              (fuel ?v - vehicle ?f - fuel-level)
              (accessible ?v - vehicle ?p1 ?p2 - location)
              (next ?f1 ?f2 - fuel-level))

 (:action drive
    :parameters (?v - vehicle ?from ?to - location
                 ?fbefore ?fafter - fuel-level)
    :precondition (and (at ?v ?from)
                       (accessible ?v ?from ?to)
                       (fuel ?v ?fbefore)
                       (next ?fbefore ?fafter))
    :effect (and (not (at ?v ?from))
                 (at ?v ?to)
                 (not (fuel ?v ?fbefore))
                 (fuel ?v ?fafter))
 )
)
```

Figure 2: A domain description in PDDL.

an unstable part of the syntax. PDDL domains are not case-sensitive, which is somewhat anachronistic in the light of standard practice in modern programming languages.

In the following sections we review the extensions made to PDDL in its development into PDDL2.1, the version of the language used in the third International Planning Competition.

## 3. Numeric Expressions, Conditions and Effects

One of the first decisions we made in the development of PDDL2.1 was to propose a definitive syntax for the expression of numeric fluents. We based our syntax on the version described in the AI Magazine article (McDermott, 2000), with some minor revisions (discussed below). Numeric expressions are constructed, using arithmetic operators, from primitive numeric expressions, which are values associated with tuples of domain objects by domain functions. Using our proposed syntax for expressing numeric assignments and updates we can express the jug-pouring operator originally described in the PDDL1.2 manual and in the AI Magazine article (see Figure 1), in PDDL2.1, as presented in Figure 4. In this example the functions capacity and amount associate the jug objects with numeric values corresponding to their capacity and current contents respectively. As can be seen in the example, we have used a prefix syntax for all arithmetic operators, including comparison predicates, in order to simplify parsing. Conditions on numeric expressions are always comparisons between pairs of numeric expressions. Effects can make use of a selection of assignment operations in order to update the values of primitive numeric expressions. These include direct assignment and relative assignments (such as increase and decrease). Numbers are not distinguished in their possible roles, so values can represent, for example, quantities of resources, accumulating utility, indices or counters.

```
(define (problem vehicle-example)
 (:domain vehicle)
 (:objects
      truck car - vehicle
      full half empty - fuel-level
      Paris Berlin Rome Madrid - location)
 (:init
      (at truck Rome)
      (at car Paris)
      (fuel truck half)
      (fuel car full)
      (next full half)
      (next half empty)
      (accessible car Paris Berlin)
      (accessible car Berlin Rome)
      (accessible car Rome Madrid)
      (acessible truck Rome Paris)
      (accessible truck Rome Berlin)
      (accessible truck Berlin Paris)
 )
 (:goal (and (at truck Paris)
             (at car Rome))
 )
)
```

Figure 3: A problem instance associated with the vehicle domain.

```
(define (domain jug-pouring)
   (:requirements :typing :fluents)
   (:types jug)
   (:functions
        (amount ?j - jug)
        (capacity ?j - jug))

   (:action pour
       :parameters (?jug1 ?jug2 - jug)
       :precondition (>= (- (capacity ?jug2) (amount ?jug2)) (amount ?jug1))
       :effect (and (assign (amount ?jug1) 0)
                    (increase (amount ?jug2) (amount ?jug1)))
)
```

Figure 4: Pouring water between jugs, PDDL2.1 style.

The differences between the PDDL2.1 syntax and the AI Magazine syntax are in the declaration of the functions and in the use of `assign` instead of `change`. We decided to only allow numeric-valued functions, making the declaration of function return types superfluous. We therefore simplified the language by requiring only the declaration of the function names and argument types, as is required for predicates. We felt that `change` was ambiguous when used alongside the operations `increase` and `decrease` and that `assign` would be clearer.

Numeric expressions are not allowed to appear as terms in the language (that is, as arguments to predicates or values of action parameters). There are two justifications for this decision — a philosophical one and a pragmatic one. Philosophically we take the view that there are only a finite number of objects in the world. Numbers do not exist as unique and independent objects in the world, but only as values of attributes of objects. Our models are object-oriented in the sense that all actions can be seen as methods that apply to the objects given as their parameters. The object-oriented view does not directly inform the syntax of our representations, but is reflected in the way in which numbers are manipulated only through their relationships with the objects that are identified and named in the initial state. Pragmatically, many current planning approaches rely on being able to instantiate action schemas prior to planning, and this is only feasible if there is a finite number of action instances. The branching of the planner's search space, at choice points corresponding to action selection, is therefore always over finite ranges. The use of numeric fluent variables conflicts with this because they could occur as arguments to any predicate and would not define finite ranges.

Our decision not to allow numbers to be used as arguments to actions rules out some actions that might seem intuitively reasonable. For example, an action to *fly* at a certain altitude might be expected to take the altitude as a number-valued argument. This is only possible in PDDL2.1 if the range of numbers that can be used is finite. From a practical point of view we think that this is unlikely to be an arduous constraint and that the benefits of keeping the logical state space finite compensates for any modelling awkwardness that results.

Functions in PDDL2.1 are restricted to be of type $Object^n \rightarrow \mathbb{R}$, for the (finite) collection of objects in a planning instance, $Object$ and finite function arity $n$. Later extensions of PDDL might introduce functions of type $Object^n \rightarrow Object$, allowing $Object$ to be extended by the application of functions to other objects. The advantage of this would be to allow objects to be referred to by their relationships to known objects. (For example, `(onTopOf ?x)` could be used to refer to the object currently on top of an object instantiating `?x`). Unfortunately, such functions present various semantic problems. In particular, the interpretation of quantified preconditions becomes significantly harder, since the collection of objects is no longer necessarily finite, so extensional interpretations are not possible. A further difficulty is the identity problem — as objects are manipulated by actions, the functional expressions that refer to them are also affected, but implicitly. For example, as objects are moved, `(onTopOf A)` can change without any action manipulating it explicitly. Managing the way in which functional terms map to specific objects in the domain (which might or might not have specific names of their own) appears to introduce considerable complication into the semantics. We believe that it is important to avoid extending PDDL with elements that are still poorly understood.

## 4. Plan Metrics

The adoption of a stable numeric extension to the PDDL core allowed us to introduce a further extension into PDDL2.1, namely a new (optional) field within the specification of problems: a plan metric. Plan metrics specify, for the benefit of the planner, the basis on which a plan will be evaluated for a particular problem. The same initial and goal states might yield entirely different optimal plans given different plan metrics. Of course, a planner might not choose to use the metric to guide its development of a solution but just to evaluate a solution *post hoc*. This approach might lead to sub-optimal, and possibly even poor quality plans, but it is a pragmatic approach to the handling of metrics which was quite widely used in the competition. This issue is discussed further in the companion paper analysing the results of the 3rd IPC in this issue (Long & Fox, 2003b).

The value `total-time` can be used to refer to the temporal span of the entire plan. Other values must all be built from primitive numeric expressions defined within a domain and manipulated by the actions of the domain. As a consequence, plan metrics can only express non-temporal metrics in PDDL2.1 domains using numeric expressions. Any arithmetic expression can be used in the specification of a metric — there is no requirement that the expression be linear. It is the domain designer's responsibility to ensure that plan metrics are well-defined (for example, do not involve divisions by zero). An example of use of a plan metric is shown in Figure 5.

The implications of having introduced this extension are far-reaching and have already helped to demonstrate some important new challenges for planning systems — particularly fully-automated systems. An enriched descriptive power for the evaluation of plans is a crucial extension for the practical use of planners, since it is almost never the case that real plans are evaluated solely by the number of actions they contain.

Metrics are described in the problem description, allowing a modeller to easily explore the effect of different metrics in the construction of solutions to problems for the same domain. In order to define a metric in terms of a specific quantity it is necessary to *instrument* that quantity in the domain description. For example, if the metric is defined in terms of overall fuel use a *fuel-use* quantity can be initialised to zero in the initial state and then updated every time fuel is consumed. In the domain shown in Figure 5 it is possible to minimise a linear combination of fuel used by each of the vehicles such as:

```
(:metric minimize (+ (* 2 (fuel-used car)) (fuel-used truck)))
```

However it is not possible to minimise distance covered since distance is not instrumented. It would be straightforward to instrument it if desired, simply by adding the appropriate initial value and incrementing effects to the domain description. Since actions cause quantities to change, instrumenting a value requires modification of the domain description itself, not just a problem file.

The use of plan metrics is subtle and can have dramatic impact on the plans being sought. Perhaps the simplest case is where all actions increase a metric that must be minimised, or decrease one that must be maximised. This is the case in the example shown in Figure 5, where any use of the drive action can only worsen the value of the plan metric (whether we use the metric shown in the figure or the maximising metric described in the last paragraph). This situation might appear to be relatively straightforward: a planner

```
(define (domain metricVehicle)
 (:requirements :strips :typing :fluents)
 (:types vehicle location)
 (:predicates (at ?v - vehicle ?p - location)
              (accessible ?v - vehicle ?p1 ?p2 - location))
 (:functions  (fuel-level ?v - vehicle)
              (fuel-used ?v - vehicle)
              (fuel-required ?p1 ?p2 - location)
              (total-fuel-used))

 (:action drive
   :parameters (?v - vehicle ?from ?to - location)
   :precondition (and (at ?v ?from)
                      (accessible ?v ?from ?to)
                      (>= (fuel-level ?v) (fuel-required ?from ?to)))
   :effect (and (not (at ?v ?from))
                (at ?v ?to)
                (decrease (fuel-level ?v) (fuel-required ?from ?to))
                (increase (total-fuel-used) (fuel-required ?from ?to))
                (increase (fuel-used ?v) (fuel-required ?from ?to)))
 )
)

(define (problem metricVehicle-example)
 (:domain metricVehicle)
 (:objects
       truck car - vehicle
       Paris Berlin Rome Madrid - location)
 (:init
       (at truck Rome)
       (at car Paris)
       (= (fuel-level truck) 100)
       (= (fuel-level car) 100)
       (accessible car Paris Berlin)
       (accessible car Berlin Rome)
       (accessible car Rome Madrid)
       (accessible truck Rome Paris)
       (accessible truck Rome Berlin)
       (accessible truck Berlin Paris)
       (= (fuel-required Paris Berlin) 40)
       (= (fuel-required Berlin Rome) 30)
       (= (fuel-required Rome Madrid) 50)
       (= (fuel-required Rome Paris) 35)
       (= (fuel-required Rome Berlin) 40)
       (= (fuel-required Berlin Paris) 40)
       (= (total-fuel-used) 0)
       (= (fuel-used car) 0)
       (= (fuel-used truck) 0)
 )
 (:goal (and (at truck Paris)
             (at car Rome))
 )
 (:metric minimize (total-fuel-used))
)
```

Figure 5: An example of a domain and problem instance describing a plan metric.

must attempt to use as few actions to solve a problem as possible. In fact, even this case is a little more complex than it appears — there can be rival plans in which one uses more actions but has lower overall cost than the other. A more complex case arises when some actions improve the quality metric while others degrade it. For example, if we use the maximising metric but also add a refuel action to the domain then driving will degrade plan quality (by reducing the fuel level of a vehicle) but refuelling will improve plan quality (by increasing the fuel level of a vehicle). In this case, a planner can attempt to use actions to improve the plan quality without those actions actually contributing to achieving the goals. For example, refuelling might not be necessary to get the vehicles to their destinations, but adding refuelling actions would improve the quality of a solution. This process could involve trading off finite and irreplaceable resources for the increased value of the plan. This would be the case if, for example, refuelling a vehicle took fuel from a finite reservoir. Alternatively a domain could allow plans of arbitrarily high value to be constructed by using more and more actions. This would occur in the metric vehicles domain using the maximising vehicle's fuel level metric if refuelling were not constrained, since the domain does not impose a limit on the fuel capacities of the vehicles.

The case in which plans are constrained by finite availability of resources, is an important and interesting form of the planning problem, but the case in which plans of arbitrarily high utility can be constructed, is obviously an ill-defined problem, since an optimal plan does not exist. It is non-trivial to determine whether a planning problem provided with a metric is ill-defined. In fact, as Helmert shows (Helmert, 2002), the introduction of numeric expressions, even in the constrained way that we have adopted in pddl2.1, makes the planning problem undecidable. The problem of finding a collection of actions which does not consume irreplaceable resources and has an overall beneficial impact on a plan metric is at least as hard as the planning problem. Therefore it is clear that determining whether a planning problem is even well-defined is undecidable. This does not make it worthless to consider planning with metrics, of course, but it demonstrates that the modelling problem, as well as the planning problem, becomes even more complex when metrics are introduced.

One strategy available to planners working with problems subject to plan metrics is to ignore the metric and simply produce a plan to satisfy the logical goals that a problem specifies. In this case, the plan quality will simply be the value, according to the metric, of the plan that happens to be constructed. This strategy is unsophisticated and it is obviously better for a planner to construct a plan guided by the specified metric. How best to use a metric to expedite the search process in a fully-automated planner is still a research issue.

## 5. Durative Actions

Most recent work on temporal planning (Smith & Weld, 1999; Bacchus & Kabanza, 2000; Do & Kambhampati, 2001) has been based on various forms of durative action. In order to facilitate participation in the competition we therefore developed two forms of durative action allowing the specification only of restricted forms of timed conditions and effects in their description. Although constrained in certain ways, these durative actions are, nevertheless, more expressive than many of the proposals previously explored, particularly in the way that they allow concurrency to be exploited. The two forms are *discretised* durative actions and *continuous* durative actions.

```
(:durative-action load-truck
    :parameters (?t - truck)
                 (?l - location)
                 (?o - cargo)
                 (?c - crane)
    :duration (= ?duration 5)
    :condition (and (at start (at ?t ?l))
                    (at start (at ?o ?l))
                    (at start (empty ?c))
                    (over all (at ?t ?l))
                    (at end (holding ?c ?o)))
    :effect (and (at end (in ?o ?t))
                 (at start (holding ?c ?o))
                 (at start (not (at ?o ?l)))
                 (at end (not (holding ?c ?o)))
)
```

Figure 6: A durative action for loading a truck. We assume no capacity constraints.

Both forms rely on a basic durative action structure consisting of the logical changes caused by application of the action. We always consider logical change to be instantaneous, therefore the continuous aspects of a continuous durative action refer only to how numeric values change over the interval of the action. Figure 6 depicts a basic durative action, *load-truck*, in which there is no numeric change.

The modelling of temporal relationships in a discretised durative action is done by means of *temporally annotated* conditions and effects. All conditions and effects of durative actions must be temporally annotated. The annotation of a condition makes explicit whether the associated proposition must hold at the *start* of the interval (the point at which the action is applied), the *end* of the interval (the point at which the final effects of the action are asserted) or over the interval from the start to the end (invariant over the duration of the action). The annotation of an effect makes explicit whether the effect is immediate (it happens at the start of the interval) or delayed (it happens at the end of the interval). No other time points are accessible, so all discrete activity takes place at the identified start and end points of the actions in the plan.

Invariant conditions in a durative action are required to hold over an interval that is open at both ends (starting and ending at the end points of the action). These are expressed using the *over all* construct seen in Figures 6 and 8. If one wants to specify that a fact *p* holds in the closed interval over the duration of a durative action, then three conditions are required: *(at start p)*, *(over all p)* and *(at end p)*.

We considered adopting the convention that *over all* constraints should apply to the start and end points as well as the open interval inside the durative action, but decided against this because it would then be impossible to express conditions that are actually only required to hold over this open interval. Examples of actions in which conditions are invariant only over the open interval include the action of loading a truck. The truck must remain at the loading location throughout the loading interval, but it can start to move away simultaneously with the loading being completed. The reason is that the start of the drive action is non-mutex with the end of the load so there is a reasonable interpretation of any plan in which driving starts at the instant that loading is completed. Actions that affect an invariant condition (such as the location of the truck) can be executed simultaneously with the end point of a durative action only if the invariant is not constrained to hold true at the end point itself. This highlights an important difference between (*over all*) and (*over all* and *at end*). If a condition is required as an end precondition as well as an invariant condition the meaning is that any action that affects the invariant must start after the end of the action requiring that invariant. For example, if we make *(at truck location)* an end precondition of the load operator as well as an invariant, the consequence is that the truck cannot drive away until after the instant at which the load has completed.

Note that, in our definition of the *load-truck* action in Figure 6, we have chosen to make the condition *(holding ?c ?o)* be a start effect and an end precondition but not an invariant condition. This means that the crane could temporarily cease to hold the cargo at some time during the interval, as long as it is holding the cargo in time to deposit it at the end of the loading interval. This makes the action quite flexible, enabling the exploitation of concurrent uses of the crane where applicable.

The *load-truck* example shows how logical change can be wrapped up into durative actions that encapsulate much of the detail involved in achieving an effect by a sequence of connected activities. Naturally it would be useful to be able to combine such actions concurrently within a plan. In the next section we consider the extent to which concurrency is allowed and the ways in which concurrent plans are interpreted.

## 5.1 The Interpretation of Concurrent plans

When time is introduced into the modelling of a domain it is possible for concurrent activity to occur in a plan. Prior to the introduction of time into PDDL all plans were interpreted as sequential — even Graphplan-concurrent plans were sequenced before being validated — so concurrency was never an issue. In PDDL2.1 plan validity can depend on exploiting concurrency correctly. Actions can overlap and co-occur, giving rise to questions over the interpretation of synchronous behaviour. We discuss the problems arising in precise synchronization in Section 10. We now explain under what constraints actions can occur concurrently within a plan involving durative actions and numeric conditions and effects.

The key difference, between durative actions in PDDL2.1 and those used by planners prior to the competition, is that we distinguish between the conditions and effects at the start and end points of the durative interval and the invariant conditions that might be specified to hold over the interval. That is, actions can have pre- and postconditions that are local to the two end-points of the action, and a planner can choose to exploit a durative action for effects it has at its start or at its end. Conditions that are invariant are distinguished from pre-conditions, enabling the exploitation of a higher degree of concurrency than is possible if preconditions are not distinguished from invariants, as in TGP (Smith & Weld, 1999), TPsys (Garrido, Onaindía, & Barber, 2001) and TP4 (Haslum & Geffner, 2001). We discuss the consequences of these design decisions, together with several examples of durative actions, in the following sections.

It is important to observe that our view of time is point-based rather than interval-based. That is, we see a period of activity in terms of intervals of state separated by time points at which state-changing activities occur. All logical state change occurs instantaneously, at the start or end point of a durative action. Propositions are true over half-open intervals that are closed on the left and open on the right. Activities might change logical state or they might update the values of numeric variables. In the discretised view of time we allow for only a finite number of activities (which we call *happenings*) between any two time points, although time itself is considered continuous and actions can be scheduled to begin at any time point.

For a plan to be considered valid, no logical condition can be both asserted and negated at the same instant. We impose the further constraint that no logical condition can both be required to hold and be asserted at the same instant. Although this might seem overly strong we claim that a plan cannot be guaranteed to be valid if the instant at which a proposition is required is exactly the instant at which it is asserted. We require that, for an action with precondition $P$ to start at time $t$, there must be a half open interval immediately preceding $t$ in which $P$ holds. This is mathematically inconsistent with $P$ being asserted at the instant at which it is required. We are conservative in our view of the validity of simultaneous update of and access to a state proposition. For example, if we have two instantaneous actions, $A$ and $B$, where $A$ has precondition $P$ and effects (*not* $P$) and $Q$, while $B$ has precondition $P \lor Q$ and effect $R$, we consider that an attempt to apply $A$ and $B$ simultaneously in a state in which $P$ holds is ill-defined. The reason is that, although $A$ switches the state from one in which $P$ holds into one in which $Q$ holds so one might suppose the precondition of $B$ to be secure, $A$ is an abstraction of a model in which the values of $P$ and $Q$ are changing and, we argue, any reliance on their values at this point of change is unstable. We adopt a rule we call *no moving targets*, by which we mean that no two actions can simultaneously make use of a value if one of the two is accessing the value to update it — the value is a moving target for the other action to access. This rule creates a behaviour for propositions in a planning state that is very much like the behaviour of variables in shared memory protected by a *mutex lock* (such as those in POSIX threads), with a difference between read and write access to the variable.

Validity also requires that no numeric value be accessed and updated simultaneously at the start or end point of a durative action. In the case of discretised durative actions, all numeric change is modelled in terms of step functions so numeric values can be accessed, or updated, during the interval of another durative action acting on that value (we provide examples in the following section) provided that any updates are consistent with all invariant properties dependent on the value. In the case of continuous durative actions, values can be simultaneously accessed and updated during the continuous process of change occurring in the interval of an action. In both the discretised and continuous cases we allow multiple simultaneous updates provided the update operations are commutative.

In order to implement the mutual exclusion relation we require *non-zero*-separation between mutually exclusive action end points. In our view, when end points are non-conflicting they can be treated as though it is possible to execute them simultaneously even though precise synchronicity cannot be achieved in the world. However, when end points are mutually exclusive the planner should buffer the co-occurrence of these points by explicitly separating them. In this way we ensure that the concurrency in the plan is at least plausible in the world.

```
(:durative-action heat-water
   :parameters (?p - pan)
   :duration (= ?duration (/ (- 100 (temperature ?p)) (heat-rate)))
   :condition (and (at start (full ?p))
                   (at start (onHeatSource ?p))
                   (at start (byPan))
                   (over all (full ?p))
                   (over all (onHeatSource ?p))
                   (over all (heating ?p))
                   (at end (byPan)))
   :effect (and   (at start (heating ?p))
                  (at end (not (heating ?p)))
                  (at end (assign (temperature ?p) 100)))
)
```

Figure 7: A simple durative action for boiling a pan of water.

Planners can exploit considerable concurrency in a domain by ensuring only that conflicting start and end points of actions are separated by a non-zero amount. A detailed specification of the mutual exclusion relation of PDDL2.1 is given in Section 8. We further discuss the implications of non-zero separation in Section 10.

## 5.2 Numeric Change within Discretised Durative Actions

This section explains how continuous change can sometimes be modelled in PDDL2.1 using durative actions with discrete effects. This is achieved by using step functions to describe instantaneous changes at the beginnings or ends of the durations of actions. Appendix A details the language constructs involved.

An example of a durative action, illustrating the use of numeric update operations, is shown in Figure 7. In this example showing a water heating action, the conditions *(full ?p)* and *(onHeatSource ?p)* must hold at the start of the interval as well as during the interval. To model this we enter these conditions as both *at start* and *over all* constraints. The action achieves as its start effect that the water is heating, and this condition is maintained invariant over the whole interval of the action. This is an example of an operator that achieves its own invariant condition, and draws attention to the fact that *over all* conditions hold over an interval that is open on the left (as well as on the right).

It should be noted that the actions in Figures 7 and 8 use fixed duration specifications. In the case of the water-boiling example this means that it is impossible to adjust the length of time over which the pan is heated and this has an impact on the context in which the action can be used. In particular, when an *assign* construct is used to update a numeric value, it is not possible for concurrent activity to affect the same value or else the model will be flawed. Because the water heating example uses an *assign* construct no concurrent activity should affect the temperature of the water. It is the responsibility of the modeller to ensure that the temperature is neither accessed nor updated during the interval over which the action is executing.

```
(:durative-action navigate
 :parameters (?x - rover ?y - waypoint ?z - waypoint)
 :duration (= ?duration (travel-time ?y ?z))
 :condition (and (at start (available ?x))
                 (at start (at ?x ?y))
                 (at start (>= (energy ?x)
                              (* (travel-time ?y ?z) (use-rate ?x))))
                 (over all (visible ?y ?z))
                 (over all (can_traverse ?x ?y ?z)))
 :effect (and (at start (decrease (energy ?x)
                              (* (travel-time ?y ?z) (use-rate ?x))))
              (at start (not (at ?x ?y)))
              (at end (at ?x ?z))))

(:durative-action recharge
 :parameters (?x - rover ?w - waypoint)
 :duration (= ?duration (recharge-period ?x))
 :condition (and (at start (at ?x ?w))
                 (at start (in-sun ?w))
                 (at start (<= (energy ?x) (capacity ?x)))
                 (over all (at ?x ?w)))
 :effect (at end (increase (energy ?x) (* ?duration (recharge-rate ?x)))))
```

Figure 8: Discretised durative actions for a rover to move between locations and to recharge.

We decided to leave it to the modeller to ensure correct behaviour of the *assign* construct because we did not want to forbid the modelling of truly discontinuous updates. For example, a durative action that models the deposit of a cheque in a bank account might have a duration of three days, with a discontinuous update to the account balance at the end of that interval — it would be inappropriate to prevent actions from accessing the balance during the three day period. In general, modelling continuous change with discrete effects is open to various pitfalls. This is the price that is paid for the convenience of not having to specify the details of the continuous processes.

The use of discretised durative actions in combination with numeric (step-function) updates requires care in modelling. In particular, it relies on the notion of *conservative resource updating*. The updating of resource levels is conservative if the consumption of a resource is modelled as if it happens at the start of a durative action, even though it actually happens continuously over the duration of the action, and production of a resource is modelled as if it happens at the end of the durative action even though, again, it might actually be produced continuously over the interval.

As an example of a discretised durative action, Figure 8 shows how the action of a rover navigating between two points is modelled. The local precondition of the start of the period is that the rover be at the start location. Local effects include that the rover consumes an appropriate amount of energy and that it is at the destination. The first of these is conservative and therefore immediate, while the second is a logical effect that occurs at the end point. This organisation ensures that no parallel activities will consume energy that has already been committed to the navigation activity. Similarly, the recharge action only makes new charge available at the conclusion of the action, so that charge gained cannot be exploited until after the recharging is complete. The use of conservative updates ensures that a model does not support invalid concurrency.

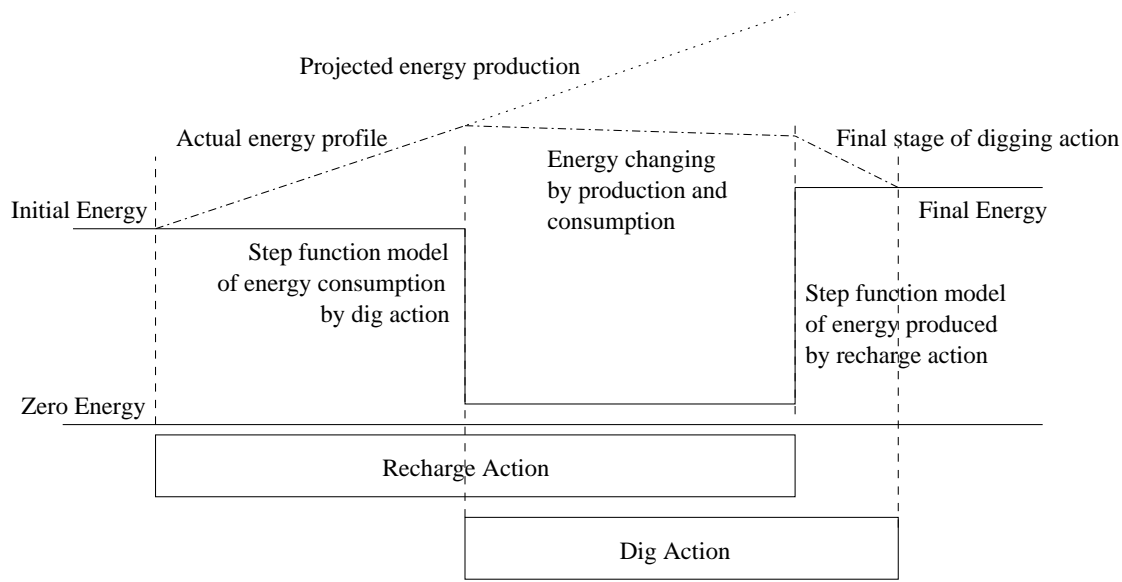

Figure 9: Using discrete actions to model the production and consumption of a resource. In reality, the recharge activity produces energy continuously and the concurrent dig activity continuously consumes it. The conservative model using step functions requires that the energy consumed by digging must be available at the start of that action, despite not having yet updated the model to show the additional energy accumulated because of the part of the recharge action so far executed. The final energy level is consistent with having used a continuous model.

Figure 9 illustrates how a recharging and a digging action (that consumes energy) would interact under a conservative energy consumption model. This model would allow concurrent actions to consume energy provided they did not consume more energy than was left under the conservative assumption that the dig action consumed all of its demands at the start and the recharge action produced nothing until the end. Note that the example assumes energy constraints but no capacity constraint.

The use of conservative updates is subtle. If there were a capacity constraint on the energy level of the rover then one would need to consider two separate resources: the energy itself and the space available for storage of energy. The dig action would consume energy at the start and only produce space at the end, while the recharge action would consume space at the start and produce charge at the end. Using this combination it would be possible to ensure that plans did not consume either resource before it was available.

Durative actions can have conditional effects. The antecedents and consequents of a conditional effect are temporally annotated so that it is possible to specify that the condition be checked at start or at end, and that the effect be asserted at either of these points. The

```
(:durative−action burnMatch                    (:action pickUp
  :parameters (?m − match ?l − location)          :parameters (?l − location ?o − object)
  :duration (and (< ?duration 5) (> ?duration 0)) :precondition (and (at ?l)
  :condition (and (at start (have ?m))                              (onFloor ?o ?l)
                  (at start (at ?l)))                               (light ?l))
  :effect (and (when (at start (dark ?l))         :effect (and (not (onFloor ?o ?l))
                     (and (at start (not (dark ?l)))              (have ?o)))
                          (at start (light ?l))))
              (at start (not (have ?m)))
              (at start (burning ?m))
              (at end (not (burning ?m)))
              (when (at start (dark ?l))
                    (and (at end (not (light ?l)))
                         (at end (dark ?l))))))
Actions
```

```
Initial state: (onFloor coin) (have aMatch) (at basement) (dark basement)
Goal: (have coin)
Problem
```

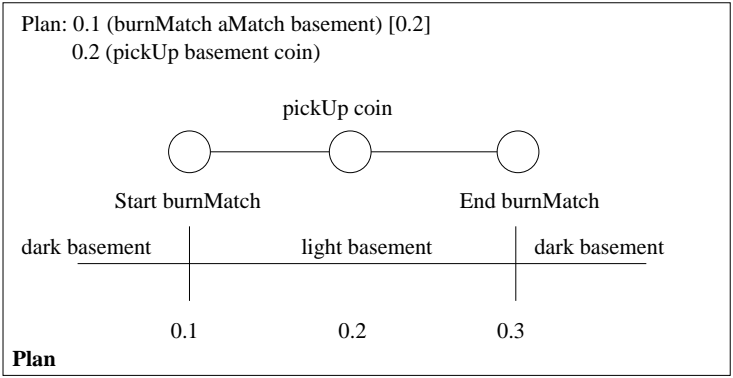

```
Plan: 0.1 (burnMatch aMatch basement) [0.2]
      0.2 (pickUp basement coin)
Plan
```

Figure 10: An example of a problem with a durative action useful for its start effects. The burning match produces the light necessary to pick up the coin.

semantics makes clear that a well-formed durative action with conditional effects cannot require the condition to be checked after the effect has been asserted. Conditional effects arise in all PDDL2.1 variants. We discuss how their occurrence in discretized durative actions is interpreted in Section 8.1.

PDDL2.1 allows the specification of duration inequalities enabling actions to be described in which external factors can be involved in determining their temporal extent. In the match-burning example shown in Figure 10 it can be seen that the effect at the start point is the only one of interest, so a planner would exploit this action for its start rather than its end effect. The duration inequality specifies that the match will burn for no longer than a specified upper bound. The model shows that the match can be put out early if the planner considers it appropriate. We discuss the use of duration inequalities further in Section 5.3.

## 5.3 Durative Actions with Continuous Effects

The objective of discrete durative actions is to abstract out continuous change and concentrate on the end points of the period over which change takes place. The syntax allows precise specification of the discrete changes at the end points of durative actions. However, when a plan needs to manage continuously changing values, as well as discretely changing ones, the durative action language and semantics need to be more powerful. General durative actions can have continuous as well as discrete effects. These increase, or decrease, some numeric variable according to a specified rate of change over time for that variable. When determining how to achieve a goal a planner must be able to access the values of these continuous quantities at arbitrary points on the time-line of the plan. We use #t to refer to the continuously changing time from the start of a durative action during its execution. For example, to express the fact that the fuel level of a plane, ?p, decreases continuously, as a function of the consumption rate of ?p, we write:

```
(decrease (fuel-level ?p) (* #t (consumption-rate ?p)))
```

This is distinctly different from:

```
(at end (decrease  (fuel-level ?p)
                   (* (flight-time ?a ?b) (consumption-rate ?p))))
```

because the latter is a single update happening at the end point of the flight action, whilst the former allows the correct calculation of the fuel level of the plane at any point in that interval. The former is a continuous effect, whilst the latter is a discrete one. Continuous effects are not temporally annotated because they can be evaluated at any time during the interval of the action. #t is local to each durative action, so that each durative action has access to a purely local "clock". Another way to interpret the expression representing continuous change is as a differential equation:

$$\frac{d}{dt}(\texttt{fuel-level ?p}) = (\texttt{consumption-rate ?p})$$

We chose to use the #t symbol instead of a differential equation because it is possible for two concurrent actions to be simultaneously modifying the same quantity. In that case, the use of differential equations would actually form an inconsistent pair of simultaneous equations, rather than having the intended effect of a combined contribution to the changing value of the quantity. Although all of the expressions describing continuous change take the form of a product of #t and some quantity, it is possible to express complex change using them with interdependent concurrent effects. For example, acceleration arises by simply increasing distance using a quantity describing velocity, while at the same time increasing velocity using a quantity describing acceleration. When dependencies between the changing terms include mutual dependencies between terms then the differential equations that arise can lead to continuous change dictated by exponential, logarithmic and exponential functions.

A plan containing continuous durative actions can assign to, consult, and continuously modify the same numeric variables concurrently (see Example 1).

In Figures 12 and 14 the discrete and continuous actions for heating a pan of water are presented (this simple model ignores heat loss). The discrete action presented in Figure 12 modifies the version presented in Figure 7 by the use of a duration *inequality* constraint.

**Example 1** *In the flying and refuelling example shown in Figure 11 it can be seen that the invariant condition, that the fuel-level be greater than (or equal to) zero during the flight, has to be maintained whilst the fuel is continuously decreasing. This could be expressed with discrete durative actions by abstracting out the continuous decrease and making the final value available at the end point of the flight. However, if a refuel operation happens during the flight time (in mid-air) then the fuel level after the flight will need to be calculated by taking into account both the continuous rate of consumption and the refuel operation. A discrete action could not calculate the fuel-level correctly because it would only have access to the distance between the source and destination of the flight, together with the rate of consumption, to determine the final fuel level. In order to calculate the fuel level correctly it is necessary to determine the time at which the refuel takes place, and to use the remaining flight-time to calculate the fuel consumed. Discrete durative actions do not give access to time points other than their own start and end points.*

*Discrete durative actions can be used to express the desired combinations of flying and refuelling by providing additional durative actions, such as fly-and-refuel, that encapsulate all of the interactions just described and end up calculating the fuel level correctly. However, this approach requires more of the domain designer than it does of the planner — the domain designer must anticipate every useful combination of behaviours and ensure that appropriate encapsulations are provided.*

*In contrast with the discrete form, the continuous action, in which the fuel consumption effect is given in terms of #t, is powerful enough to express the fact that the mid-flight refuelling of the plane affects the final fuel level in a way consistent with maintaining the invariant of the fly action.*

```
(:durative-action fly
     :parameters (?p - airplane ?a ?b - airport)
     :duration (= ?duration (flight-time ?a ?b))
     :condition (and (at start (at ?p ?a))
                     (over all (inflight ?p))
                     (over all (>= (fuel-level ?p) 0)))
     :effect (and (at start (not (at ?p ?a)))
                  (at start (inflight ?p))
                  (at end (not (inflight ?p)))
                  (at end (at ?p ?b))
                  (decrease (fuel-level ?p)
                            (* #t (fuel-consumption-rate ?p)))))))

(:action midair-refuel
     :parameters (?p)
     :precondition (inflight ?p)
     :effect (assign (fuel-level ?p) (fuel-capacity ?p)))
```

Figure 11: A continuous durative action for flying.

```
(:durative-action heat-water
    :parameters (?p - pan)
    :duration (at end (<= ?duration (/ (- 100 (temperature ?p))
                                                    (heat-rate))))
    :condition (and (at start (full ?p))
                    (at start (onHeatSource ?p))
                    (at start (byPan))
                    (over all (full ?p))
                    (over all (onHeatSource ?p))
                    (over all (heating ?p))
                    (at end (byPan)))
    :effect (and    (at start (heating ?p))
                    (at end (not (heating ?p)))
                    (at end (increase (temperature ?p)
                                (* ?duration (heat-rate)))))))
)
```

Figure 12: A discrete durative action for heating a pan of water, using a variable duration.

Duration inequalities add significant expressive power over duration equalities. Duration constraints that express inequalities are associated with an additional requirements flag because of the extended expressiveness over fixed-duration discrete durative actions.

In both actions, the logical post-condition of the start of the period is that the pan is heating. The conditions that the pan be heating, full and on the heat source are invariant, although the presence of the agent (by the pan) is only a local precondition of the two end-points and is not invariant. In the first action the duration is modelled by expressing the following duration inequality constraint:

```
(at end (<= ?duration (/ (- 100 (temperature ?p)) (heat-rate))))
```

and the effect at the end-point of the discrete durative action is that the temperature of the pan is increased by `(* ?duration (heat-rate))` (where `heat-rate` is a domain constant). In the continuous action of Figure 14 the duration constraint is unnecessary since the invariant

```
(over all (<= (temperature ?p) 100))
```

is added to ensure that the pan never exceeds boiling.

The durative action in Figure 12 models the heating pan in the face of possible concurrent activities affecting the temperature. The duration inequality allows the planner to adapt the duration to take account of other temperature-affecting activity in a way that is not possible when the duration is specified using an equality constraint. The duration constraint ensures that the temperature never exceeds boiling by checking, as a precondition for the updating activity, that the computed temperature increase can be executed without exceeding the boiling point. If this temperature increase would exceed boiling the plan is invalid. The temperature at the end of the interval of execution is computed from the current temperature and the heating rate, together with the duration over which the heating action has been active (see further discussion in Example 2).

> **Example 2** *If a plan attempts to further heat the pan (say by applying a blowtorch to the pan), during the heat-water interval then, provided that the concurrent action ends before the end of the heat-water action, the duration constraint will be seen to have been violated if the duration has been chosen so that the overall increase in temperature would exceed boiling. If the concurrent activity ends simultaneously with the heat-water action then the no-moving-targets rule would be violated because the duration constraint would attempt to access the temperature at the same time point as the concurrent action attempted to update it.*
>
> *Figure 13 depicts these two situations. In this figure, apply-blowtorch is a durative action that applies heat to an object (in this case, the pan). In part (a) of the figure the duration constraint will be violated if the duration of the heat-water action is sufficient to cause the temperature to increase beyond boiling when combined with the heat increase caused by the blowtorch — in that case that the plan will be invalid. The planner can choose a value for duration that avoids this violation. In part (b) the plan will be determined invalid regardless of the duration of the action because of the no-moving-targets rule. Notice that this model does not attempt to model the consequences of continued heating of the pan after the boiling point, so plans with actions that cause this to occur are simply invalid. However, PDDL2.1 can be used to model more of the physical situation, so that the consequences are explicit and the planner can choose to exploit them or avoid them accordingly.*

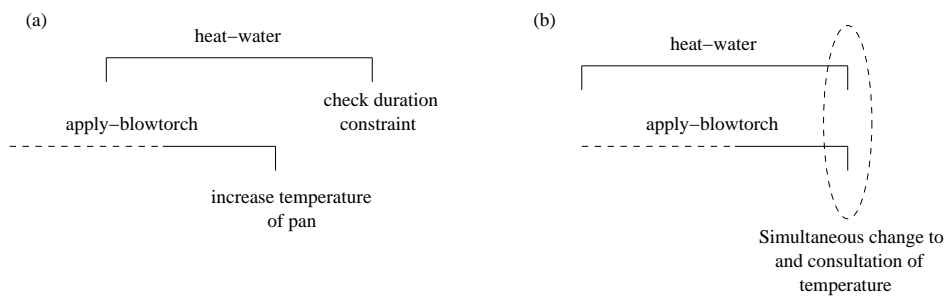

Figure 13: Heating a pan with a discrete durative action, concurrently with another heating activity.

```
(:durative-action heat-water
   :parameters (?p - pan)
   :duration ()
   :condition (and (at start (full ?p))
                   (at start (onHeatSource ?p))
                   (at start (byPan))
                   (over all (full ?p))
                   (over all (onHeatSource ?p))
                   (over all (heating ?p))
                   (over all (<= (temperature ?p) 100))
                   (at end (byPan)))
   :effect (and    (at start (heating ?p))
                   (at end (not (heating ?p)))
                   (increase (temperature ?p) (* #t (heat-rate))))
)
```

Figure 14: A continuous durative action for heating a pan of water.

The use of duration inequalities adds significant expressive power even when using discrete durative actions. For example, the plan depicted in part (a) of Figure 13, which illustrates the use of the water-heating action shown in Figure 12 while concurrently heating the pan with a blowtorch, will be considered valid provided that there is a duration value that satisfies the duration constraint in the water-heating action. This brings us very close to the expressive power available with continuous durative actions because it gives the planner the power to exploit concurrent interacting activities enacting changes on the same numeric valued variable (see Example 3). Attempting to express continuous change using only duration inequalities does not give precisely equivalent behaviour, because the discretisation forces actions that access changing numeric values to be separated, by some small temporal interval, from the actions that change those values in order to resolve their mutual exclusion. In a continuous model this is not necessary because the true value of a numeric variable is available for consultation at any time during the continuous process of change.

In the discrete semantics presented in Section 8 we exploit the fact that the only changes that can occur when a plan is executed are at points corresponding to the times of happenings, so the plan can be checked by looking at the activity focussed in this finite happening sequence. In fact, provided continuous effects are restricted to linear functions of time with only first order effects (which requires that no continuous effects can affect numeric expressions contributing to the rate of change of another numeric valued variable), and invariants are restricted to linear functions of changing quantities, it is still possible to restrict attention to the happening sequence even when using continuous actions.

Non-linear effects and higher-order rates of change create difficulties since it is possible for an invariant to be satisfied at the end points of an interval, without having necessarily been satisfied throughout the interval. In these cases it is no longer sufficient to insert invariant checking actions at fixed mid-points in the happening sequence of a plan in order to validate its behaviour. However, provided that effects are first-order and linear, and invariants are linear in continuously changing values, then, despite the fact that arbitrary

---

**Example 3** *It is possible with discrete durative actions, with duration inequalities, to model the effects of adding an egg to the heating water when the water is at, say, 90 degrees. We do this by applying two heat-water actions, around an add-egg action, in such a way that the overall duration of the two heat-water actions is exactly the duration required to boil the water from its original temperature. However, the way the heat-water action is currently modelled means that the heat will be turned off before the egg is added, and then turned on again to complete the heating, since the temperature is only updated when the durative action terminates. With continuous durative actions the egg can be added whilst the single heat-water action is in progress since the temperature of the pan is continuously updated. So, discrete durative actions with duration inequalities allow us to approximate continuous activity by appending a finite sequence of discrete intervals in an appropriate way. The no moving targets rule means that the end points of these intervals will be separated by non-zero, arbitrarily small, time gaps. This is not required when using continuous actions because, in contrast to the step-function effects of discrete actions, continuous effects are not localised at a single point.*

---

time points within action intervals are accessible to the planner, it is only necessary to gain access to numeric values at the start- and end-points of the actions in the plan that refer to them, together with finitely many mid-points for invariant-checking actions. The values are not required at all other points. This is so because continuous durative actions do not support the modelling of exogenous events, so it is not necessary to take into account the exogenous activity of the environment in determining the validity of a plan.

### 5.4 Related Approaches

Time is an important numerically varying quantity. The simplest way to reason about time is to adopt a *black box* durative action model in which change happens at the ends of their durative intervals. This is the approach taken in the language used by TGP (Smith & Weld, 1999), for example, in which durative actions encapsulate continuous change so that the correct values of any affected variables are guaranteed only at the end points of the implied intervals. All of the logical and numeric effects of a durative action are enacted at the end of the action and are undefined during the interval of its execution. All undeleted preconditions must remain true throughout the interval. There is no syntactic distinction between *pre*conditions and *invariant* conditions in this action representation. A simplistic way of ensuring correct action application is to prevent concurrent actions that refer to the same facts, but this excludes many intuitively valid plans.

A more sophisticated approach allows preconditions to be annotated with time points, or intervals, so that the requirement that a condition be true at some point, or over some interval, within the duration of the action can be expressed. This is the approach taken in Sapa (Do & Kambhampati, 2001). For example, using such an annotated precondition it would be possible to express the requirement that some chemical additive be added within two minutes of the start of a tank-filling action. If effects can also be specified to occur at arbitrary points within the duration of the action then it is possible to express effects that occur before the end of the specified duration. It is also possible to distinguish between conditions that are local to specific points in the duration of the action and those that are invariant throughout the action.

Allowing reference to finitely many time points between the start and end of actions makes the language more complex without adding to its expressive power. Where time points are strictly scheduled relative to the start of the action the effect can be achieved through the use of a sequence of linked durative actions. We decided to keep PDDL2.1 simple by restricting access to only the end points of actions.

In TLPlan (Bacchus & Ady, 2001) a similar, but more constrained, approach is adopted in which actions are applied instantaneously but can have delayed effects. The delays for effects can be arbitrary and different for each effect. However, invariants cannot be specified because the preconditions are checked at the instant of application and subsequent delayed effects are separated from the action which initiated them.

Several planners have been developed to use networks of temporal constraints (Ghallab & Laruelle, 1994; Jonsson, Morris, Muscettola, & Rajan, 2000; El-Kholy & Richards, 1996) to handle temporal structure in planning problems. Efficient algorithms exist for handling such constraints (Dechter, Meiri, & Pearl, 1991) which make them practical for managing large networks. The domain models constructed using PDDL2.1 certainly lend themselves to treatment by similar techniques, but are not constrained to be handled in this way.

## 6. Introduction to the Semantics of PDDL2.1

In Sections 7 and 8 we provide a formal semantics for the numeric extension and temporal extension of PDDL2.1. Together these sections contain 20 definitions. The lengthy treatment is necessary because the semantics we have developed adds four significant extensions over classical planning and the semantics Lifschitz developed for STRIPS (Lifschitz, 1986). These are:

- the introduction of time, so that plans describe behaviour relative to a real time line;

- related to the first extension, the treatment of concurrency — actions can be executed in parallel, which can lead to plans that contain concurrent interacting processes (although these processes are encapsulated in durative actions in PDDL2.1);

- an extension to handle numeric-valued fluents;

- the use of conditional effects, both alone and in conjunction with all of the above extensions.

The semantics is built on a familiar state-transition model. The requirements of the semantics can be reduced to four essential elements.

1. To define what is a state. The introduction of both time and numeric values complicate the usual definition of a state as a set of atoms.

2. To define when a state satisfies a propositional formula representing a goal condition or precondition of an action. An extension of the usual interpretation of a state as a valuation in which an atom is true if and only if the atom is in the state (the Closed World Assumption) is required in order to handle the numeric values in the state.

3. To define the state transition induced by application of an action. The update rule for the logical state must be supplemented with an explanation of the consequences for the numeric part of the state.

4. To define when two actions can be applied concurrently and how their concurrent application affects the application of those actions individually.

The structure of the definitions is as follows. Definitions 1 to 15, given in Section 7, define what it means for a plan to be valid when the plan consists of only non-durative actions. Definitions 1 to 6 set up the basic terminology, the foundational structures and the framework for handling conditional effects and primitive numeric expressions. Definition 2 meets the first requirement identified above, defining states. Definition 9 meets the second requirement, defining when a goal description is satisfied in a state. Definition 11 defines a simple plan, extending the classical notion of a sequence of actions by adding time. Definitions 12 meets the fourth requirement, by defining when two actions cannot be executed concurrently. Definition 13 meets the third requirement, defining what we mean by execution of actions, including concurrent execution of actions. Definitions 14 and 15 define the execution of a plan and what it means for a plan to be valid, given the basis laid in the previous definitions.

In Section 8 the semantics is extended to give meaning to durative actions. We begin with Definition 16, which defines ground durative actions analogously to Definition 6 for simple (that is, non-durative) actions. Similarly, Definition 17 parallels the definition of a simple plan (Definition 11) and Definitions 19 and 20 parallel those for the execution and validity of simple plans (Definitions 14 and 15). Definition 18 is the critical definition for the semantics of plans with durative actions, supplying a transformation of temporal plans into simple plans, whose validity according to the semantics of purely simple plans, can be used to determine the validity of the original temporally structured plans.

## 7. The Semantics of Simple Plans

The semantics we define in this section extends the essential core of Lifschitz' STRIPS semantics (1986) to handle temporally situated actions, possibly occurring simultaneously, with numeric and conditional effects.

**Definition 1 Simple Planning Instance** *A* simple planning instance *is defined to be a pair*

$$I = (Dom, Prob)$$

*where $Dom = (Fs, Rs, As, arity)$ is a 4-tuple consisting of (finite sets of) function symbols, relation symbols, actions (non-durative), and a function arity mapping all of these symbols to their respective arities. $Prob = (Os, Init, G)$ is a triple consisting of the objects in the domain, the initial state specification and the goal state specification.*

*The* primitive numeric expressions *of a planning instance, $PNEs$, are the terms constructed from the function symbols of the domain applied to (an appropriate number of) objects drawn from $Os$. The* dimension *of the planning instance, dim, is the number of distinct primitive numeric expressions that can be constructed in the instance.*

The atoms *of the planning instance, Atms, are the (finitely many) expressions formed by applying the relation symbols in Rs to the objects in Os (respecting arities).*

*Init consists of two parts: $Init_{logical}$ is a set of literals formed from the atoms in Atms. $Init_{numeric}$ is a set of propositions asserting the initial values of a subset of the primitive numeric expressions of the domain. These assertions each assign to a single primitive numeric expression a constant real value. The goal condition is a proposition that can include both atoms formed from the relation symbols and objects of the planning instance and numeric propositions between primitive numeric expressions and numbers.*

*As is a collection of action schemas (non-durative actions) each expressed in the syntax of* PDDL. *The primitive numeric expression schemas and atom schemas used in these action schemas are formed from the function symbols and relation symbols (used with appropriate arities) defined in the domain applied to objects in Os and the schema variables.*

The semantics shows how instantiated action schemas can be interpreted as state transitions, in a similar way to the familiar state transition semantics defined by Lifschitz. An important difference is that states can no longer be seen as simply sets of propositions, but must also account for the numeric expressions appearing in the planning instance and the time at which the state holds. This is achieved by extending the notion of state.

**Definition 2 Logical States and States** *Given the finite collection of atoms for a planning instance I, $Atms_I$, a logical state is a subset of $Atms_I$. For a planning instance with dimension dim, a state is a tuple in $(\mathbb{R}, \mathbb{P}(Atms_I), \mathbb{R}_\perp^{dim})$ where $\mathbb{R}_\perp = \mathbb{R} \cup \{\perp\}$ and $\perp$ denotes the undefined value. The first value is the time of the state, the second is the logical state and the third value is the vector of the dim values of the dim primitive numeric expressions in the planning instance.*

*The initial state for a planning instance is $(0, Init_{logical}, \mathbf{x})$ where $\mathbf{x}$ is the vector of values in $\mathbb{R}_\perp$ corresponding to the initial assignments given by $Init_{numeric}$ (treating unspecified values as $\perp$).*

Undefined values are included in the numeric ranges because there are domains in which some terms start undefined but can nevertheless be initialised and exploited by actions.

To interpret actions as state transition functions it is necessary to achieve two steps. Firstly, since (in PDDL2.1) plans are only ever constructed from fully instantiated action schemas, the process by which instantiation affects the constructs of an action schema must be defined and, secondly, the machinery that links primitive numeric expressions to elements of the vector of real values in a state and that allows interpretation of the numeric updating behaviours in action effects must be defined. Since the mechanisms that support the second of these steps also affect the process in the first, the treatment of numeric effects is described first.

**Definition 3 Assignment Proposition** *The syntactic form of a numeric effect consists of an assignment operator (`assign`, `increase`, `decrease`, `scale-up` or `scale-down`), one primitive numeric expression, referred to as the lvalue, and a numeric expression (which is an arithmetic expression whose terms are numbers and primitive numeric expressions), referred to as the rvalue.*

*The* assignment proposition *corresponding to a numeric effect is formed by replacing the assignment operator with its equivalent arithmetic operation (that is* (`increase p q`) *becomes* (`= p (+ p q)`) *and so on) and then annotating the lvalue with a "prime".*

*A numeric effect in which the assignment operator is either* `increase` *or* `decrease` *is called an* additive assignment effect*, one in which the operator is either* `scale-up` *or* `scale-down` *is called a* scaling assignment effect *and all others are called* simple assignment effects.

A numeric effect defines a function of the numeric values in the state to which an action is applied determining the value of a primitive numeric expression in the resulting state. For the convenience of a uniform treatment of numeric expressions appearing in pre- and post-conditions, we transform the functions into propositions that assert the equality of the post-condition value and the expression that is intended to define it. That is, rather than writing an effect (`increase p q`) as a function $f(p) = p + q$, we write it as the proposition (`= p' (+ p q)`). The "priming" distinguishes the postcondition value of a primitive numeric expression from its precondition value (a convention commonly adopted in describing state transition effects on numeric values). The binding of the primitive numeric expressions to their values in states is defined in the following definition.

**Definition 4 Normalisation** *Let I be a planning instance of dimension $dim_I$ and let*

$$index_I \ : \ PNEs_I \ \rightarrow \ \{1, \dots, dim\}$$

*be an (instance-dependent) correspondence between the primitive numeric expressions and integer indices into the elements of a vector of $dim_I$ real values, $\mathbb{R}_{\perp}^{dim_I}$.*

*The normalised form of a ground proposition, p, in I is defined to be the result of substituting for each primitive numeric expression $f$ in p, the literal $X_{index_I(f)}$. The normalised form of p will be referred to as $\mathcal{N}(p)$. Numeric effects are normalised by first converting them into assignment propositions. Primed primitive numeric expressions are replaced with their corresponding primed literals. $\mathbf{X}$ is used to represent the vector $\langle X_1 \dots X_n \rangle$.*

In Definition 4, the replacement of primitive numeric expressions with indexed literals allows convenient and consistent substitution of the vector of actual parameters for the vector of literals $\mathbf{X}$ appearing in a state.

With the machinery supporting treatment of numeric expressions complete, it is now possible to consider the process of instantiating action schemas. This process is managed in two steps. The first step is to remove constructs that we treat as syntactic sugar in the definition of a domain. These are conditional effects and quantified formulae. We handle both of these by direct syntactic transformations of each action schema into a set of action schemas considered to be equivalent. The transformation is similar to that described by Gazen and Knoblock (1997). Although it would be possible to give a semantic interpretation of the application of conditional effects directly, the transformation allows us to significantly simplify the question of what actions can be performed concurrently.

**Definition 5 Flattening Actions** *Given a planning instance, I, containing an action schema $A \in As_I$, the set of action schemas $flatten(A)$, is defined to be the set S, initially containing A and constructed as follows:*

- While $S$ contains an action schema, $X$, with a conditional effect, (when P Q), create two new schemas which are copies of $X$, but without this conditional effect, and conjoin the condition P to the precondition of one copy and Q to the effects of that copy, and conjoin (not P) to the precondition of the other copy. Add the modified copies to $S$.

- While $S$ contains an action schema, $X$, with a formula containing a quantifier, replace $X$ with a version in which the quantified formula ( Q ( $var_1 \ldots var_k$ ) P) in $X$ is replaced with the conjunction (if the quantifier, $Q$, is forall) or disjunction (if $Q$ is exists) of the propositions formed by substituting objects in $I$ for each variable in $var_1 \ldots var_k$ in P in all possible ways.

These steps are repeated until neither step is applicable.

Once flattened, actions can be grounded by the usual substitution of objects for parameters:

**Definition 6 Ground Action** *Given a planning instance, $I$, containing an action schema $A \in As_I$, the set of ground actions for $A$, $GA_A$, is defined to be the set of all the structures, $a$, formed by substituting objects for each of the schema variables in each schema, $X$, in flatten($A$) where the components of $a$ are:*

- Name *is the name from the action schema, $X$, together with the values substituted for the parameters of $X$ in forming $a$.*

- $\text{Pre}_a$, *the precondition of $a$, is the propositional precondition of $a$. The set of ground atoms that appear in $\text{Pre}_a$ is referred to as $\text{GPre}_a$.*

- $\text{Add}_a$, *the positive postcondition of $a$, is the set of ground atoms that are asserted as positive literals in the effect of $a$.*

- $\text{Del}_a$, *the negative postcondition of $a$, is the set of ground atoms that are asserted as negative literals in the effect of $a$.*

- $\text{NP}_a$, *the numeric postcondition of $a$, is the set of all assignment propositions corresponding to the numeric effects of $a$.*

*The following sets of primitive numeric expressions are defined for each ground action, $a \in GA_A$:*

- $L_a = \{f | f \text{ appears as an lvalue in } a\}$

- $R_a = \{f | f \text{ is a primitive numeric expression in an rvalue in } a \text{ or appears in } Pre_a\}$

- $L_a^* = \{f | f \text{ appears as an lvalue in an additive assignment effect in } a\}$

Some comment is appropriate on the last definition: an action precondition might be considered to have two parts — its logical part and its numeric expression-dependent part. Unfortunately, these can be interdependent. For example:

```
(or (clear ?x) (>= (room-in ?y) (space-for ?z)))
```

might be a precondition of an action. In order to handle such conditions, we need to check whether they are satisfied given not only the current logical state, but also the current values of the domain numeric expressions. The inclusion of a numeric component in the state makes it necessary to ensure the correct substitution of the numeric values for the expressions used in the action precondition. This is achieved using the normalisation process from Definition 4 in Definition 9. In contrast, the postcondition of an action cannot contain interlocked numeric and logical effects, so it *is* possible to separate the effects into the distinct numeric and logical components.

**Definition 7 Valid Ground Action** *Let a be a ground action. a is* valid *if no primitive numeric expression appears as an lvalue in more than one simple assignment effect, or in more than one different type of assignment effect.*

Definition 7 ensures that an action does not attempt inconsistent updates on a numeric value. Unlike logical effects of an action which cannot conflict, it is possible to write a syntactic definition of an action in which the effects are inconsistent, for example by assigning two different values to the same primitive numeric expression.

**Definition 8 Updating Function** *Let a be a valid ground action. The* updating function *for a is the composition of the set of functions:*

$$\{\text{NPF}_p \ : \ \mathbb{R}_\perp^{dim} \to \ \mathbb{R}_\perp^{dim} \,|\, p \in NP_a\}$$

*such that* $\text{NPF}_p(\mathbf{x}) = \mathbf{x}'$ *where for each primitive numeric expression* $x'_i$ *that does not appear as an lvalue in* $\mathcal{N}(p)$, $x'_i = x_i$ *and* $\mathcal{N}(p)[\mathbf{X}' := \mathbf{x}', \mathbf{X} := \mathbf{x}]$ *is satisfied.*

*The notation* $\mathcal{N}(p)[\mathbf{X}' := \mathbf{x}', \mathbf{X} := \mathbf{x}]$ *should be read as the result of normalising p and then substituting the vector of actual values* $\mathbf{x}'$ *for the parameters* $\mathbf{X}'$ *and actual values* $\mathbf{x}$ *for formal parameters* $\mathbf{X}$.

Definition 8 defines the function describing the update effects of an action. The function ensures that all of the reals in the vector describing the numeric state remain unchanged if they are not affected by the action (this is the numeric-state equivalent of the persistence achieved for propositions by the STRIPS assumption). For other values in the vector, the normalisation process is used to substitute the correctly indexed vector elements for the primitive numeric expressions appearing as lvalues (which are the primed vector elements corresponding to values in the post-action state) and rvalues (the unprimed values appearing in the pre-action state). The tests that must be satisfied in order to ensure correct behaviour of the functions in the composition simply confirm that the arithmetic on the rvalues is correctly applied to arrive at the lvalues. The requirement that the action be valid ensures that the composition of the functions in Definition 8 is well-defined, since all of the functions in the set commute, so the composition can be carried out in any order.

The various sets of primitive numeric expressions defined in the Definition 6 allow us to conveniently express the conditions under which two concurrent actions might interfere with one another. In particular, we are concerned not to allow concurrent assignment to the same primitive numeric expression, or concurrent assignment and inspection. We *do* allow concurrent increase or decrease of a primitive numeric expression. To allow this we will have to apply collections of concurrent updating functions to the primitive numeric expressions. This can be allowed provided that the functions commute. Additive assignments do commute, but other updating operations cannot be guaranteed to do so, except if they do not affect the same primitive numeric expressions or rely on primitive numeric expressions that are affected by other concurrent assignment propositions. It would be possible to make a similar exception for scaling effects, but additive assignment effects have a particularly important role in durative actions that is not shared by scaling effects, so for simplicity we allow concurrent updates only with these effects. We use the three sets of primitive numeric expressions to determine whether we are in a safe situation or not. Within a single action it is possible for the rvalues and lvalues to intersect. That is, an action can update primitive numeric expressions using current values of primitive numeric expressions that are also updated by the same action. All rvalues will have the values they take in the state prior to execution and all lvalues will supply the new values for the state that follows.

**Definition 9 Satisfaction of Propositions** *Given a logical state, $s$, a ground propositional formula of* PDDL2.1, $p$, *defines a predicate on* $\mathbb{R}^{dim}_\perp$, $Num(s, p)$, *as follows:*

$$Num(s, p)(\mathbf{x}) \quad iff \quad s \models \mathcal{N}(p)[\mathbf{X} := \mathbf{x}]$$

*where $s \models q$ means that $q$ is true under the interpretation in which each atom, $a$, that is not a numeric comparison, is assigned true iff $a \in s$, each numeric comparison is interpreted using standard equality and ordering for reals and logical connectives are given their usual interpretations. $p$ is satisfied in a state $(t, s, \mathbf{X})$ if $Num(s, p)(\mathbf{X})$.*

*Comparisons involving $\perp$, including direct equality between two $\perp$ values are all undefined, so that enclosing propositions are also undefined and not satisfied in any state.*

**Definition 10 Applicability of an Action** *Let $a$ be a ground action. $a$ is* applicable *in a state $s$ if the $Pre_a$ is satisfied in $s$.*

## 7.1 Semantics of a Simple Plan

A simple plan, in PDDL2.1, is a sequence of timed actions, where a timed action has the following syntactic form:

$$t : (action\, p_1 \ldots p_n)$$

In this notation $t$ is a positive rational number in floating point syntax and the expression $(action\, p_1 \ldots p_n)$ is the name and actual parameters of the action to be executed at that point in time. In more complex plans simple and durative actions, with or without numeric-valued effects or preconditions, can co-occur — the semantics of such plans is discussed in Section 8. No special separators are required to separate timed actions in the sequence and the actions are not required to be presented in time-sorted order. It is possible for multiple actions to be given the same time stamp, indicating that they should be executed concurrently. It should be emphasised that the earliest point at which activity occurs within a plan must be strictly after time 0. This constraint follows from the decision to make the initial state be the state existing at time 0, together with the decision, in the semantics, that actions have their effects in the interval that is *closed on the left*, starting at the time when the action is applied, while preconditions are tested in the interval that is *open on the right* that precedes the action.

In order to retain compatibility with the output of current planners the following concession is made: if the plan is presented as a sequence of actions with no time points, then it is inferred that the first action is applied at time 1 and the succeeding actions apply in sequence at integral time points one unit apart.

A simple plan is a slight generalisation of the more familiar STRIPS-style classical plan, since actions are labelled with the time at which they are to be executed.

**Definition 11 Simple Plan** *A* simple plan*, $SP$, for a planning instance, $I$, consists of a finite collection of* timed simple actions *which are pairs $(t, a)$, where $t$ is a rational-valued time and $a$ is an action name.*

*The* happening sequence*, $\{t_i\}_{i=0\ldots k}$ for $SP$ is the ordered sequence of times in the set of times appearing in the timed simple actions in $SP$. All $t_i$ must be greater than 0. It is possible for the sequence to be empty (an empty plan).*

*The* happening *at time $t$, $E_t$, where $t$ is in the happening sequence of $SP$, is the set of (simple) action names that appear in timed simple actions associated with the time $t$ in $SP$.*

A plan thus consists of a sequence of happenings, each being a set of action names applied concurrently at a specific time, the sequence being ordered in time. The times at which these happenings occur forms the happening sequence. It should be noted that action names are ambiguous when action schemas contain conditional effects — the consequence of flattening is to have split these actions into multiple actions with identical names, differentiated by their preconditions. However, at most one of each set of actions with identical names can be applicable in a given logical state, since the precondition of each action in such a set will necessarily be inconsistent with the precondition of any other action in the set, due to the way in which the conditional effects are distributed between the pairs of action schemas they induce.

In order to handle concurrent actions we need to define the situations in which the effects of those actions are consistent with one another. This issue was first discussed in Section 5.1. The mutual exclusion rule for PDDL2.1 is an extension of the idea of *action mutex* conditions for GraphPlan (Blum & Furst, 1995). The extension handles two extra features: the extended expressive power of the language (to include arbitrary propositional connectives) and the introduction of numeric expressions. We make a very conservative condition for actions to be executed concurrently, which ensures that there is no possibility of interaction. This rules out cases where intuition might suppose that concurrency is possible. For example, the actions:

```
(:action a
  :precondition (or p q)
  :effect (r))

(:action b
  :precondition (p)
  :effect (and (not p) (s)))
```

could, one might suppose, be executed simultaneously in a state in which both $p$ and $q$ hold. The following definition asserts, however, that the two actions are mutex. The reason we have chosen such a constrained definition is because checking for mutex actions must be tractable and handling the case implied by this example would appear to require checking the consequence of interleaving preconditions and effects in all possible orderings. The condition on primitive numeric expressions has already been discussed — it determines that the update effects can be executed concurrently and that they do not affect values which are then tested by preconditions (regardless of whether the results of those tests matter to the satisfaction of their enclosing proposition). This is the rule of *no moving targets*: no concurrent actions can affect the parts of the state relevant to the precondition tests of other actions in the set, regardless of whether those effects might be harmful or not. It might be considered odd that the preconditions of one action cannot refer to literals in the add effects of a concurrent action. We require this because preconditions can be negative, in which case their interaction with add effects is analogous to the interaction between positive preconditions and delete effects. The no moving targets rule makes the cost of determining whether a set of actions can be applied concurrently polynomial in the size of the set of actions and their pre- and post-conditions.

**Definition 12 Mutex Actions** *Two grounded actions, a and b are* non-interfering *if*

$$GPre_a \cap (Add_b \cup Del_b) = GPre_b \cap (Add_a \cup Del_a) = \emptyset$$
$$Add_a \cap Del_b = Add_b \cap Del_a = \emptyset$$
$$L_a \cap R_b = R_a \cap L_b = \emptyset$$
$$L_a \cap L_b \subseteq L_a^* \cup L_b^*$$

*If two actions are not non-interfering they are* mutex.

The last clause of this definition asserts that concurrent actions can only update the same values if they both do so by additive assignment effects.

We are now ready to define the conditions under which a simple plan is valid. We can separate the executability of a plan from whether it actually achieves the intended goal. We will say that a plan is valid if it is executable *and* achieves the final goal. Executability is defined in terms of the sequence of states that the plan induces by sequentially executing the happenings that it defines.

**Definition 13 Happening Execution** *Given a state, $(t, s, \mathbf{x})$ and a happening, $H$, the* activity *for $H$ is the set of grounded actions*

$$A_H = \{a | the\ name\ for\ a\ is\ in\ H,\ a\ is\ valid\ and\ Pre_a\ is\ satisfied\ in\ (t, s, \mathbf{x})\}$$

*The* result *of executing a happening, $H$, associated with time $t_H$, in a state $(t, s, \mathbf{x})$ is* undefined *if $|A_H| \neq |H|$ or if any pair of actions in $A_H$ is mutex. Otherwise, it is the state $(t_H, s', \mathbf{x}')$ where*

$$s' = (s \setminus \bigcup_{a \in A_H} Del_a) \cup \bigcup_{a \in A_H} Add_a$$

*and $\mathbf{x}'$ is the result of applying the composition of the functions $\{\mathrm{NPF}_a \mid a \in A_H\}$ to $\mathbf{x}$.*

Since the functions $\{NPF_a \mid a \in A_H\}$ must affect different primitive numeric expressions, except where they represent additive assignment effects, these functions will commute and therefore the order in which the functions are applied is irrelevant. Therefore, the value of $\mathbf{x}'$ is well-defined in this last definition. The requirement that the activity for a happening must have the same number of elements as the happening itself is simply a constraint that ensures that each action name in the happening leads to a valid action that is applicable in the appropriate state. We have already seen that conditional effects induce the construction of families of grounded actions, but that at most one of each family can be applicable in a state. If none of them is applicable for a given name, then this must mean that the precondition is unsatisfied, regardless of the conditional effects. In this case, we are asserting that the attempt to apply the action has undefined interpretation.

**Definition 14 Executability** *A simple plan, $SP$, for a planning instance, $I$, is executable if it defines a happening sequence, $\{t_i\}_{i=0...k}$, and there is a sequence of states, $\{S_i\}_{i=0...k+1}$ such that $S_0$ is the initial state for the planning instance and for each $i = 0 \ldots k$, $S_{i+1}$ is the result of executing the happening at time $t_i$ in $SP$.*

*The state $S_{k+1}$ is called the* final state *produced by $SP$ and the state sequence $\{S_i\}_{i=0...k+1}$ is called the* trace *of $SP$. Note that an executable plan produces a unique trace.*

**Definition 15 Validity of a Simple Plan** *A simple plan (for a planning instance, $I$) is valid if it is executable and produces a final state $S$, such that the goal specification for $I$ is satisfied in $S$.*

## 8. The Semantics of Durative Actions

Plans with durative actions with discrete effects can be given a semantics in terms of the semantics of simple plans. Handling durative actions that have continuous effects is more complex — we discuss this further in Section 9.

Durative actions appearing in a plan must be given with an additional field indicating the duration. This is given with the syntax:

$$t : (action\, p_1 \ldots p_n)\, [d]$$

where $d$ is a rational valued duration, written in floating point syntax.

Durative actions are introduced into the framework we have defined so far by generalising Definition 1 to include durative action schemas. The definition of the grounded action must now be extended to define the form of grounded durative actions. However, this definition can be given in such a way that we associate with each durative action two simple (non-durative) actions, corresponding to the end points of the durative action. These simple actions can, together, simulate almost all of the behaviour of the durative action. The only aspects that are not captured in this pair of simple actions are the duration of the durative action and the invariants that must hold over that duration. These two elements can, however, be simply handled in a minor extension to the semantics of simple plans, and this is the approach we adopt. By taking this route we avoid any difficulties in establishing the effects of interactions between durative actions — this is all handled by the semantics for the concurrent activity within a simple plan. As we will see, one difficulty in this account is the handling of durative actions with conditional effects that contain conditions and effects that are associated with different times or conditions that must hold over the entire duration of the action. Since these cases complicate the semantics we will postpone treatment of them until the next section and begin with durative actions without conditional effects.

The mapping from durative actions to non-durative actions has the important consequence that the mutex relation implied between non-durative actions is (advantageously) weaker than the strong mutex relation used in, for example, TGP (Smith & Weld, 1999). Two durative actions can be applied concurrently provided that the end-points of one action do not interact either with the end-points (if simultaneous) or the invariants of the other action.

**Definition 16 Grounded Durative Actions** *Durative actions are grounded in the same way as simple actions (see Definition 6), by replacing their formal parameters with constants from the planning instance and expanding quantified propositions. The definition of durative actions requires that the condition be a conjunction of temporally annotated propositions. Each temporally annotated proposition is of the form* (at start p), (at end p) *or* (over all p), *where p is an unannotated proposition. Similarly, the effects of a durative action (without continuous or conditional effects) are a conjunction of temporally annotated simple effects.*

*The duration field of DA defines a conjunction of propositions that can be separated into $DC_{start}^{DA}$ and $DC_{end}^{DA}$, the duration conditions for the start and end of DA, with terms being arithmetic expressions and* ?duration. *The separation is conducted in the obvious way, placing* at start *conditions into $DC_{start}^{DA}$ and* at end *conditions into $DC_{end}^{DA}$.*

*Each grounded durative action, DA, with no continuous effects and no conditional effects defines two parameterised simple actions $DA_{start}$ and $DA_{end}$, where the parameter is the* ?duration *value, and a single additional simple action $DA_{inv}$, as follows.*

*$DA_{start}$ ($DA_{end}$) has precondition equal to the conjunction of the set of all propositions, p, such that* (at start p) *(*(at end p)*) is a condition of DA, together with $DC_{start}^{DA}$ ($DC_{end}^{DA}$), and effect equal to the conjunction of all the simple effects, e, such that* (at start e) *(*(at end e)*) is an effect of DA (respectively).*

*$DA_{inv}$, is defined to be the simple action with precondition equal to the conjunction of all propositions, p, such that* (over all p) *is a condition of DA. It has an empty effect.*

*Every conjunct in the condition of DA contributes to the precondition of precisely one of $DA_{start}$, $DA_{end}$ or $DA_{inv}$. Every conjunct in the effect of DA contributes to the effect of precisely one of $DA_{start}$ or $DA_{end}$. For convenience, $DA_{start}$ ($DA_{end}$, $DA_{inv}$) will be used to refer to both the entire (respective) simple action and also to just its name.*

The actions $DA_{start}$ and $DA_{end}$ are parameterised by ?duration and this parameter must be substituted with the correct duration value in order to arrive at the two simple actions corresponding to the start and end of a durative action.

**Definition 17 Plans** *A plan, P, with durative actions, for a planning instance, I, consists of a finite collection of* timed actions *which are pairs, each either of the form $(t, a)$, where t is a rational-valued time and a is a simple action name – an action schema name together with the constants instantiating the arguments of the schema, or of the form $(t, a[t'])$, where t is a rational-valued time, a is a durative action name and $t'$ is a non-negative rational-valued duration.*

**Definition 18 Induced Simple Plan** *If $P$ is a plan then the* happening sequence *for $P$ is $\{t_i\}_{i=0...k}$, the ordered sequence of time points formed from the set of times*[1]

$$\{t \mid (t,a) \in P \; or \, (t, a[t']) \in P \; or \, (t - t', a[t']) \in P\}$$

*The* induced simple plan *for a plan $P$, simplify$(P)$, is the set of pairs defined as follows:*

- $(t, a)$ *for each $(t, a) \in P$ where $a$ is a simple (non-durative) action name.*

- $(t, a_{start}[?\mathbf{duration} := t'])$ *and $(t + t', a_{end}[?\mathbf{duration} := t'])$ (these expressions are simple timed actions – the square brackets denote substitution of $t'$ for* `?duration` *in this case) for all pairs $(t, a[t']) \in P$, where $a$ is a durative action name.*

- $((t_i + t_{i+1})/2, a_{inv})$ *for each pair $(t, a[t']) \in P$ and for each $i$ such that $t \le t_i < t + t'$, where $t_i$ and $t_{i+1}$ are in the happening sequence for $P$.*

The process of transforming a plan into a simple plan involves introducing actions to represent the end points of the intervals over which the durative actions in the plan are applicable. Duration constraints convert into simple preconditions on start or end actions, requiring the substitution of a numeric value for the `?duration` field to complete the conversion into simple actions. The complication to this process is that invariants cannot be associated with the end points, but must be checked throughout the interval. This is achieved by adding to the simple plan a collection of special actions responsible for checking the invariants. These actions are added between each pair of happenings in the original plan lying between the start and end point of the durative action. Because the semantics of simple plans requires that the preconditions of actions in the plan be satisfied, even though they might have no effects, the consequence of putting these monitoring actions into the simple plan is to ensure that the original plan is judged valid only if the invariants remain true, firstly, after the start of the durative action and, subsequently, after each happening that occurs throughout the duration of the durative action. One possibility is to make these monitoring actions occur at the same times as the updating actions, but this would require values to be accessed at the same time as they might be being updated, violating the no moving targets rule. In order to avoid this problem the monitoring actions are interleaved with the updating actions by inserting them midway between pairs of successive happenings in the interval over which each durative action is executed. Only happenings in the original plan need be considered when carrying out this insertion, since the invariant-checking actions themselves cannot have any effect on the states in which they are checked.

Alternative treatments of invariants are possible, but an important advantage of the approach we have taken is that the semantics rests, finally, on a state-transition model in a form that is familiar to the planning community. That is, plans can be seen as recipes for state-transition sequences, with each state-transition being a function from the current state of the world to the next. However, durative actions complicate this picture because they rely on a commitment, once a durative action has been started, to follow it through to completion. That commitment involves some sort of communication across the duration of the plan. The communication can be managed by structures outside the plan, that examine the trace, or by artificial modification of the plan itself to ensure that states carry extra information from the start to the end of the durative action. The latter approach has the disadvantage that as durative actions become more complex the artificial components that must be added to the plan become more intrusive. This is particularly apparent in the treatment of conditional effects that require conditions tested at the start of a durative action, or across its duration, but effects that are triggered at the end, since these cases require some sort of "memory" in the state to remember the status of the tested conditions from the start of the durative action to the end point. These memory conditions allow us to avoid embedding an entire execution history in a state by substituting an *ad hoc* memory of the history for just those propositions and at just those times it is required. The management of conditional effects of this form, in the mapping from durative actions to simple actions, is discussed further in Section 8.1.

---

1. Care should be taken in reading this definition — the last disjunct allows the time corresponding to the *end* of execution of a durative action to be included as a happening time.

We can now conclude the definitions supporting the validity of a plan with durative actions.

**Definition 19 Executability of a Plan** *A plan, P (for a planning instance), is executable if the induced simple plan for P, simplify(P) is executable, producing the trace* $\{S_i = (t_i, s_i, \mathbf{v}_i)\}_{i=0...k}$.

**Definition 20 Validity of a Plan** *A plan, P (for a planning instance), is valid if it is executable and if the goal specification is satisfied in the final state produced by its induced simple plan.*

## 8.1 Durative Actions with Conditional Effects

We now explain how the mapping described in the previous section is extended to deal with durative actions containing conditional effects.

First, we observe that temporally annotated conditions and effects can be accumulated, because the temporal annotation distributes through logical conjunction. Therefore, we can convert conditional effects so that their conditions are simple conjunctions of at most one *at start* condition, at most one *at end* condition and at most one *over all* condition. It should be noted that we do not allow logical connectives other than conjunction in combining temporally annotated propositions. Allowing other connectives would create significant further complexity in the semantics and create potentially paradoxical opportunities for communication from future states to earlier states. Similarly to conditions, durative action effects can be reduced to a conjunction of at most one *at start* effect and at most one *at end* effect. Treatment of conditional effects then divides into three cases. The first case is very straightforward: any effect in a durative action of the form (`when (at t p) (at t q)`), where the condition and the effect bear the same single temporal annotation, can be transformed into a simple conditional effect of the form (`when p q`) attached to the start or end simple action according to whether $t$ is *start* or *end*. Since this case is straightforward we will not explicitly extend the previous definitions to cope with it. The second case is one in which the condition of a condition effect has *at start* conditions and the effect has *at end* effects.

Note that we consider conditional effects in which the effects occur at the start, but with conditions dependent on the state at the end or over the duration of the action, to be meaningless. This is because they reverse the expected behaviour of causality, where cause precedes effect. In any attempt to validate a plan by constructing a trace such reversed causality would be a huge problem, since we could not determine the initial effects of applying a durative action until we had seen what conditions held over the subsequent interval and conclusion of its activity, but, equally, we could not see what the effects of activity during the interval would be without seeing the initial effects of applying the durative action. This paradox is created by the opportunity for an action to change the past.

To handle this second case we need to modify the state after the start of the durative action to "remember" whether the start conditions were satisfied and communicate this to the end of the durative action where it can then be simply looked up in the (then) current state to determine whether the conditional effect should be applied. We apply a transformation to conditional effects of the form (when (and (at start ps) (at end pe)) (at end q)) into a conditional effect added to the start simple action, (when ps ($M_{ps}$)), and a conditional effect added to the end simple action, (when (and pe ($M_{ps}$)) q), where $M_{ps}$ is a special new proposition, unique to the particular conditional effect of the particular application of the durative action being transformed. By ensuring that this proposition is unique in this way, there is no possibility of any other action in the plan interfering with it, so it represents an isolated memory of the fact that ps held in the state at which the durative action was started. If a conditional effect does not have *at end* conditions, the same transformation can be applied, simply ignoring pe in the previous discussion. Figure 15 depicts the transformation of a single durative action, $A$, with a conditional effect, into a collection of level 2 actions, complete with the appropriate "memory" proposition (in this case called $P^*$).

The importance of the memory introduced in this transformation is explained in Figures 16 and 17. Figure 16 shows the ambiguity that results from not remembering how a state, on the trajectory of a plan, was reached. The figure illustrates that if one is in a state $(P, Q, \neg R)$ at the point when durative action $A$ (as described in Figure 15) ends, it is impossible to determine from the state alone whether $R$ should be added or not. This is because it is possible to have reached the state $(P, Q, \neg R)$ by at least two different paths, with at least one path having seen $A$ started in a state in which $P$ held and at least one path having seen $A$ started in a state in which $P$ did not hold (using an action, achieve-$P$, with $P$ as its only effect). The state $(P, Q, \neg R)$ does not contain any information to disambiguate which path was used to reach it, and hence cannot determine the correct value of $R$ after $A$ ends.

The third, and final, case is where the durative action has conditional effects of the form:

```
(when (and (at start ps) (over all pi) (at end pe)) (at end q)).
```

Again, if the effect has no *at start* or *at end* conditions the following transformation can be applied simply ignoring ps or pe as appropriate. In this case we need to construct a transformation that "remembers" not only whether ps held in the state at which the durative action is first applied, but also whether pi holds throughout the interval from the start to the end of the durative action. Unlike the invariants of durative actions, these conditions are not *required* to hold for the plan to be valid, but only determine what effects will occur at the end of the durative action. The idea is to use intervening monitoring actions, rather as we did for invariants in definition 18. This is achieved by adding a further effect to the start action: `(M`$_{pi}$`)`. Then, the monitoring (simple) actions that are required have no precondition, but a single conditional effect: `(when (and (M`$_{pi}$`) (not pi)) (not (M`$_{pi}$`)))`. Once again, `M`$_{pi}$ is a special new proposition unique to the conditional effect for the application instance of the durative action being transformed. The monitoring actions are added at all the intermediate points that are used for the monitoring actions in Definition 18. The same transformation used in the second case above is required again for the *at start* condition, ps, so `(when ps (M`$_{ps}$`))` is added as a conditional effect to the start simple action. Finally, we add a conditional effect to the end (simple) action: `(when (and (M`$_{ps}$`) (M`$_{pi}$`) pe) q)`. The effect of this machinery is to ensure that if the proposition `pi` becomes false at any time between the start and end of the durative action then `M`$_{pi}$ will be deleted, but otherwise at the end of the durative action `M`$_{ps}$ will hold precisely if `ps` held at the start of the action and `M`$_{pi}$ will hold precisely if `pi` has held over the entire duration of the durative action. Therefore, the conditional effect of the end action achieves the intuitively correct behaviour of asserting its conditional effect precisely when the *at start* condition held at the start of the durative action, the *at end* condition holds at the end of the durative action and its *over all* condition has held throughout the duration of the action.

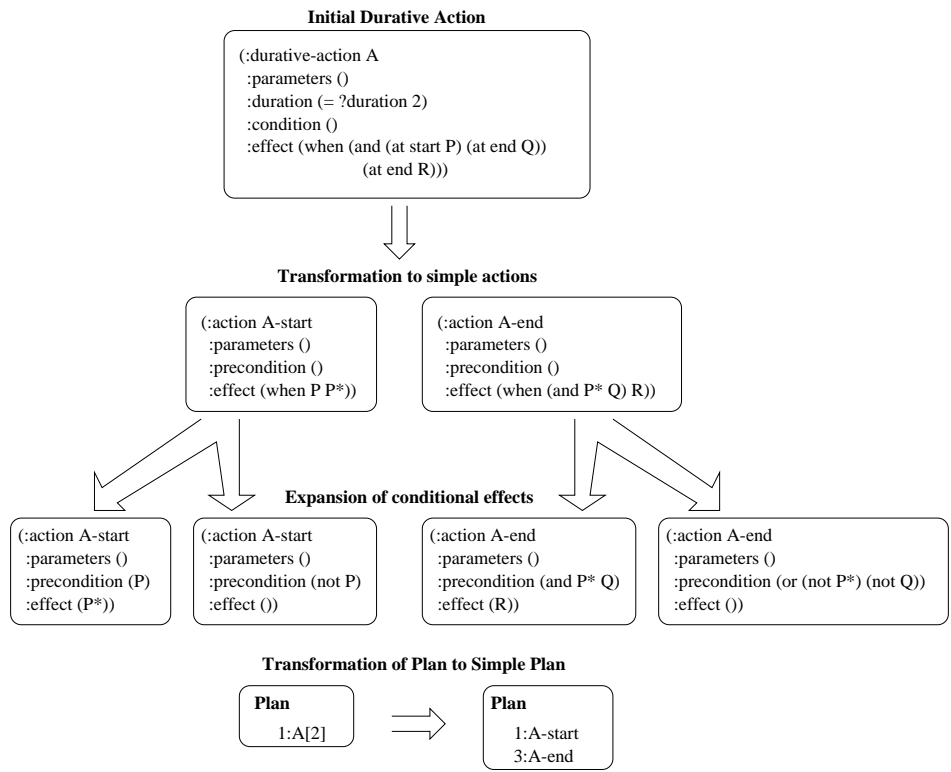

Figure 15: Conversion of a durative action into non-durative actions and their grounded forms.

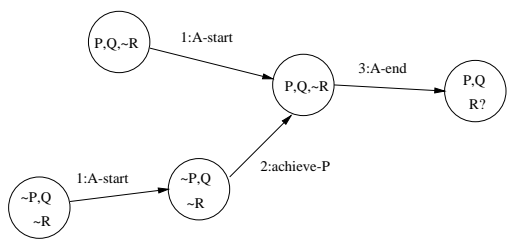

Figure 16: Flawed state space resulting from failure to record the path traversed when conditional effects span the interval of a durative action. The arc labelled *achieve-P* indicates the possible application of some action that achieves the proposition P.

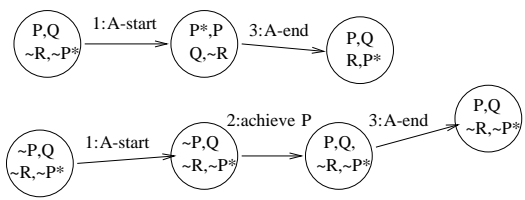

Figure 17: Correct state space showing use of "memory" proposition P*. The arc labelled *achieve P* indicates the possible application of some action that achieves the proposition P.

The addition of these new memory-checking actions means that it is no longer true to claim that the added actions cannot change the state. However, memory propositions are unique to the task of communication for a single action instance, so the effects that memory-checking actions might have on these have no implications for other invariants.

## 9. The Semantics of Continuous Durative Actions

The introduction of continuous durative actions complicates the semantics. It is no longer possible to handle invariants by insertion of simple actions between other happenings in a plan to test their continued satisfaction. In fact, continuous effects can, in principle, cause an invariant to be satisfied over some parts of an interval and not over others. Ignoring invariants for a moment, updates to numeric values caused by continuous effects can be applied as discrete updates at time points within the interval over which they apply. These updates behave slightly differently to the discrete updates we have seen in durative actions with discrete effects, since it *is* possible for a continuous update to affect a variable that is concurrently affected by a discrete update, or examined by a precondition, without creating an inconsistency. For example, if the water heating action in Figure 14 is applied with the concurrent addition of an egg to the pan with a precondition that the temperature of the water is between 90 and 95 degrees then the value of the temperature can be examined at the moment of application of the action adding the egg. This is because the temperature change is actually happening over the interval between the start of the heating and the point at which the egg is added, rather than as a discrete update at the point the egg is added. The temperature is not actually changed at the instant of the addition of the egg.

In this section we summarise the semantics for continuous actions. Where the semantics for discrete durative actions is defined in terms of the familiar state-transition semantics, the continuous semantics introduces a different formulation.

**Definition 21 A Continuous Durative Action** *A* continuous effect *is an effect expression that includes the symbol* #t. *A continuous durative action* *is a durative action with at least one continuous effect.*

**Definition 22 Continuous Update Function** *Let $C$ be a set of ground continuous effects for a planning instance, $I$, and $St = (t, S, \mathbf{X})$ be a state. The* continuous update function *defined by $C$ for state $St$ is the function $f_C : \mathbb{R} \to \mathbb{R}^n$, where $n$ is $dim_I$, such that:*

$$\frac{df_C}{dt} = g$$

*and*

$$f_C(0) = \mathbf{X}$$

*where $g$ is the update function generated for an action $a$ with:*

$$NP_a = \{ \, (\texttt{<op>} \ \texttt{P Q}) \, | \, (\texttt{<op>} \ \texttt{P} \ (\texttt{*} \ \texttt{\#t Q})) \in C \}$$

Definition 22 shows how the continuous effects of several continuous durative actions can be combined to create a single system of simultaneous differential equations whose solution, given an appropriate starting point, defines the evolution of the continuously varying values.

**Definition 23 Induced Continuous Plan** *Let $I$ be a planning instance that includes continuous durative actions and $P$ be a plan for $I$. The* induced continuous plan *for $P$ is a triple, $(S, Invs, Cts)$, where $S$ is $simplify(P)$, $Invs$ is the set of* invariant constraints:

$$Invs = \{(Q, t, t+d) \, | \, (t, a[d]) \in P \text{ and } (\texttt{over all Q}) \text{ is an invariant for } A\}$$

*Let $t_i$ and $t_{i+1}$ be two consecutive times in the happening sequence for $simplify(P)$. The set of active continuous effects over $(t_i, t_{i+1})$ is:*

$$\{Q \mid (t, a[d]) \in P, (t_i, t_{i+1}) \subseteq [t, t+d] \text{ and } Q \text{ is a continuous effect of } a\}$$

*and $Cts$ is the set of systems of continuous effects:*

$$Cts = \{(C, t_i, t_{i+1}) \mid C \text{ is the set of active continuous effects over } (t_i, t_{i+1})\}$$

The components of a continuous plan separate out the invariant conditions and continuous effects from the rest of the simple plan in order to allow correct application of the continuous updates and to allow confirmation that the invariants hold in the face of the continuous effects.

**Definition 24 Trace** *Let $I$ be a planning instance that includes continuous durative actions, $P$ be a plan for $I$, $(SP, Inv, Cts)$ be the induced continuous plan for $P$, $\{t_i\}_{i=0...k}$ be the happening sequence for $S$ and $S_0$ be the initial state for $I$. The trace for $P$ is the sequence of states $\{S_i\}_{i=0...k+1}$ defined as follows:*

- *If there is no element $(C, t_i, t_{i+1}) \in Cts$ then $S_{i+1}$ is the state resulting from applying the happening at $t_i$ in the simple plan $SP$ to the state $S_i$.*

- *If $(C, t_i, t_{i+1}) \in Cts$ then let $T_i$ be the the state formed by substituting $f(t_{i+1} - t_i)$ for the numeric part of state $S_i$, where $f$ is the continuous update function defined by $C$ for state $S_i$. Then $S_{i+1}$ is the state resulting from applying the happening at $t_i$ in the simple plan $SP$ to the state $T_i$. If $f$ is undefined for any element in $Cts$ then so is the trace.*

Definition 24 defines a trace in a similar fashion to the traces for simple plans and plans with durative actions. The key difference is the need to apply the continuous updates. These are handled by solving the systems of simultaneous differential equations across each interval in which they are active and then applying the result to update the numeric values across that interval. Of course, this is easier to describe than it is to do, since solving arbitrary simultaneous differential equations algorithmically is not generally possible. Under certain constraints this semantics can be implemented in order to confirm the validity of a plan automatically.

**Definition 25 Invariant Safe** *Let $I$ be a planning instance that includes continuous durative actions, $P$ be a plan for $I$, $(S, Inv, Cts)$ be the induced continuous plan for $P$ and $\{S_i\}_{i=0...k+1}$ be the trace for $P$. For each $(C, t_i, t_{i+1}) \in Cts$ let $f_i$ be the continuous update function defined by $C$ for $S_i$. $P$ is invariant safe if, for each $f_i$ that is defined and for each $(Q, t, u) \in Inv$ such that $[(t_i, t_{i+1})] = I \subseteq (t, u)$, then $\forall x \in I$, $\mathbb{N}um(s, Q)(f_i(x))$ where $s$ is the logical state in $S_i$.*

*In this definition, the symbols $[(..)]$ are used to mean that the interval $I$ can be closed or open at either end.*

From a semantic point of view, invariants must be checked at every point in the interval over which they apply. When the interval contains only finitely many discrete changes then the obligation can be met by considering only the finite number of points at which change occurs (a fact that is exploited for discrete durative action plan semantics in Definition 18). When there is continuous change the obligation is much harder to meet. In practice, the invariants can be checked by examining the possible roots of the function describing continuous change, but finding those roots can be very difficult in general. Again, suitable constraints on the forms of differential equations expressed in a domain can make the validation problem tractable.

The last two definitions simply assemble the components to arrive at analogous definitions to those for executability and validity of simple plans and plans with durative actions.

**Definition 26 Executability of a Plan** *A plan $P$ containing continuous durative actions, for planning instance $I$, with induced continuous plan $(S, Invs, Cts)$. $P$ is* executable *if the trace for $P$ is defined, $\{S_i\}_{i=0...k+1}$, and it is invariant safe.*

**Definition 27 Plan Validity** *A plan $P$ containing durative actions, for planning instance $I$ is* valid *if it is executable, with trace $\{S_i\}_{i=0...k+1}$ and $S_{k+1}$ satisfies the goal in $I$.*

## 10. Plan Validation

Plan validation is an important part of the use of PDDL, particularly in its role for the competition. With approximately 5000 plans to consider in the competition in 2002, it can be seen that automation is essential. The validation problem is tractable for propositional versions of PDDL because plans are finite and can be validated simply by simulation of their execution. The issue is more complicated for PDDL2.1 because the potential for concurrent activity, possibly in the face of numeric change, makes it necessary to ensure that invariant properties are protected and that concurrent activity is non-interfering.

When durative actions are used there is a question over whether a plan should be considered valid if it does not contain all of the end points of the actions initiated in the plan. When an action is exploited in a plan for the effect it has at the end of its duration it is clear that that end point will be present in the plan, but when an action was selected for its start effect this is less clear. A match-striking action is performed for its start effect, not in order to have a burned out match at the end of a brief interval. It could be argued that, having obtained the desired start effect the end of the action is irrelevant and the plan can terminate (as soon as all goals are achieved) without ensuring that all initiated actions end safely. Indeed, the plan search process in Sapa (Do & Kambhampati, 2001) can terminate whilst there are still queued events awaiting the advancement of time. However, it is possible to conceive of situations in which the end point of an action, incorporated only for its start effect, introduces inconsistencies in the plan so that its inclusion would make the plan invalid. In these cases it seems that plan validity could be compromised by ignoring end effects.

In order to avoid having to resolve these complexities, we have taken the view that a PDDL2.1 plan is valid only if all action start and end points are explicit within the plan. Having identified that this is the case we then proceed to confirm that all happenings within the plan are mutex-free.

Plan validation is decidable for domains including discretized and, under certain constraints, continuous durative actions because all activity is encapsulated with the durative actions explicitly identified by a plan. This makes the trace induced by a plan finite and hence checkable. We therefore observe that the validation problem for PDDL2.1 is decidable even when actions contain duration inequalities. This is because the work in determining how the duration inequalities should be solved has already been completed in the finished plan so validation of the plan can proceed by simulation of its execution, as is the case for PDDL plans. The problem is tractable for domains without continuous effects, but the introduction of continuous effects can, in principle, allow expression of domains with very complex functions describing continuous numeric change (Howey & Long, 2002). Under the assumption that continuous effects are restricted to description in terms of simple linear or quadratic functions, without any interactions between concurrent continuous effects, plan validity is tractable. The cost in practice is increased however, since it may be necessary to solve polynomials in order to check invariants. Validation of plans containing more complex expressions of change is being explored.

Although plan validity checking is tractable, there is a subtlety that arises because of the need to represent plans syntactically and the difficulties involved in expressing numbers with arbitrary precision. In principle, all of the values that are required to describe valid plans are algebraic (assuming we constrain continuous effects as indicated above), and therefore finitely representable. In practice, expecting planners to handle numbers as algebraic expressions seems unnecessarily complicated and it is far more reasonable to assume that numbers will be represented as finite precision floating point values. Indeed, the syntax we have adopted for the expression of plans restricts planners to expressing times as finite precision floating point values. With this constraint, and because of limitations on the precision of floating point computations in implementations of plan validation systems, it is necessary to take a more pragmatic view of the validation process and accept that numeric conditions will have to be evaluated to a certain tolerance. Otherwise, it can occur that there is no way to report a plan to the necessary degree of accuracy for it to have a valid interpretation under the semantics we defined in Section 8. In most cases, a plan that specifies time points to, for example, four significant digits, is a reasonable abstraction of the execution time activity that will be needed to control the flow system. No plan can specify time points absolutely precisely, so abstraction is forced upon the planner by the fact that it is working with models of the world and not the physical world itself. The problem, then, is one of the relationship between the theoretical semantics and the pragmatic concerns of automated validation.

In Figure 18 this relationship is depicted in terms of what kinds of plans can be automatically validated. The left side of the picture describes the theoretical semantics, with the arrow indicating the link between plans and their interpretation under the theoretical semantics. For example, it is possible to construct a domain and problem for which a plan that requires an action to happen at time $\sqrt{2}$ is a meaningful semantic object, but for which a plan that specifies that the action happen at time 1.41 is not a meaningful semantic object because 1.41 is not equal to $\sqrt{2}$. These two plans are distinct, and only one is correct (under the assumed constraints). The right side of the picture depicts the pragmatic validation of syntactic plan objects. The two control plans, though distinct in the semantics, can map to the same syntactic object if we assume that the validation is subject to a tolerance of 0.01.

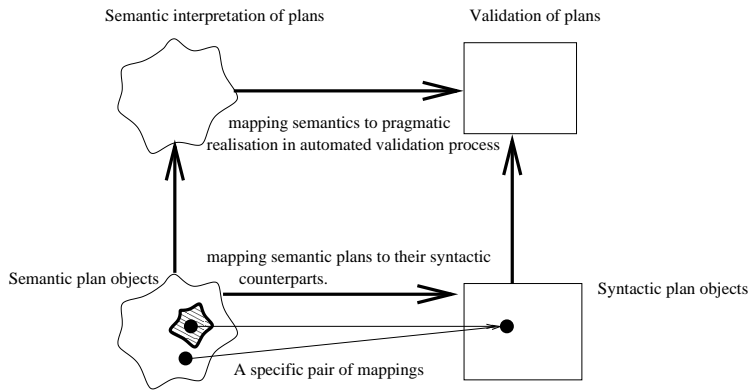

Figure 18: The pragmatic mapping between semantics of plans and their validation by automated computational processes. The shaded area contains plans that cannot be interpreted within the theoretical semantics. It can be seen that a plan in this collection is indistinguishable from a meaningful plan when mapped to the syntactic side of the picture.

These plans both map to the syntactic object in which 1.41 approximates the value $\sqrt{2}$. This syntactic plan can be validated using the pragmatic validation processes necessary for automatic validation of describable syntactic plans, which will check for validity subject to the tolerance of 0.01. The pragmatic constraints on the representations of plans, the expectations about representations of numeric values in planners and validators and their consequences are all reasonable assumptions given that the models against which we check validity are, in any case, abstractions, at some non-zero tolerance, of the world. In practice, it is a problem to accept plans at specified tolerance levels only in pathological cases, while arithmetic precision in computer representations of floats has an immediate and negative impact if one tries to take the stronger line that plans should only be accepted if they are strictly valid according to the formally precise evaluation of expressions.

Finally, there is an interesting philosophical issue that arises and is discussed by Henzinger and his co-authors (1997, 2000). It is, in fact, not possible to achieve exact precision in the measurement of time or other continuous numeric quantities. Henzinger *et al.* have considered this problem through the development of *robust automata*. Robust automata only accept a trace if there exists a *tube* of traces within a distance $\epsilon > 0$ of the original trace, all of which are acceptable by the original acceptance criteria. These are called *fuzzy tubes* indicating that time is fuzzily, rather than precisely, detectable. This idea offers a path to a formal semantics that is closer to defining plans that are robust to the imprecision in an executive's ability to measure time. Unfortunately, checking fuzzy tubes is intractable. We currently compromise by adopting an $\epsilon$ value, used as the tolerance in checking that numeric values fulfil numeric constraints during plan execution, to also represent the minimum separation of conflicting end points within plans. This is consistent with the idea that if the planner assumes that an executive is willing to abstract to the indicated tolerance level in the checking of preconditions for actions then it is unreasonable to suppose that a plan can make use of finer grained measurements in determining when actions should be applied. At the moment the value of $\epsilon$ is set in the validation process, and only communicated informally to planner-engineers, but it might be better to allow a domain designer to define an $\epsilon$ appropriate for use in the particular domain. There remain several issues concerning the correct management of the buffers during validation (particularly the usual problem concerning the transitivity of "fuzzy closeness") which are important issues for temporal reasoning as a whole and are not restricted to the planning context. We do not yet have solutions to all of these problems.

## 11. Related Work: Representing and Reasoning about Time

Representation of, and reasoning with, statements about time and the temporal extent of propositions has long been a subject of research in AI including planning research (Allen, 1984; McDermott, 1982; Sandewall, 1994; Kowalski & Sergot, 1986; Laborie & Ghallab, 1995; Muscettola, 1994; Bacchus & Kabanza, 2000). Important issues raised during the extension of PDDL to handle temporal features have, of course, already been examined by other researchers, for example in Shanahan's work (1990) on continuous change within the event calculus, in Shoham's work (1985) and Reichgelt's work (1989) on temporal reasoning and work on non-reified temporal systems (Bacchus, Tenenberg, & Koomen, 1991). Vila (1994) provides an excellent survey of work in temporal reasoning in AI. In this section we briefly review some of the central issues that have been addressed, and their treatment in the literature, and set PDDL2.1 in the context of research in temporal logics.

Several researchers in temporal logics have considered the problems of reasoning about concurrency, continuous change and temporal extent. These works have focussed on the problem of reasoning about change when the world is described using arbitrary logical formulae, and most have been concerned with making meta-level statements (such as that effect cannot precede cause). The need to handle complex logical formulae makes the frame problem difficult to resolve, and an approach based on circumscription (McCarthy, 1980) and default reasoning (Reiter, 1980) is typical. The STRIPS assumption provides a simple solution to the frame problem when states are described using atomic formulae. The classical planning assumption is that states can be described atomically but this is not a general view of the modelling of change. Although simplifying, this assumption is surprisingly expressive. The bench mark domains introduced in the third International Planning Competition suggest that atomic modelling is powerful enough to capture some complex domains which closely approximate real problems. The temporal reasoning issues we confront are not simplified as a consequence of having made a simplifying assumption about how states are updated. We remain concerned with the major issues of temporal reasoning: concurrency, continuous change and temporal extent.

In the development of PDDL2.1 we made a basic decision to consider the end points of durative actions as instantaneous state transitions. This allows us to concentrate on the truth of propositions at points instead of over intervals. The decision to consider actions in this way is similar to that made by many temporal reasoning researchers (Shanahan, 1990; McCarthy & Hayes, 1969; McDermott, 1982). In the context of PDDL2.1 the approach has the advantage of smoothly integrating with the classical planning view of actions as state transitions. Nevertheless, Allen has shown that a temporal ontology based on intervals can be a basis for planning (Allen, 1984, 1991) and several planning systems have been strongly influenced by the intervals approach (Muscettola, 1994; Rabideau, Knight, Chien, Fukunaga, & Govindjee, 1999). Allen later moved away from his initial position that instants are not required, introducing the notion of *moments* (Hayes & Allen, 1987), which are a concept that attempts to reconcile the stance that nothing is instantaneous (so there should only be intervals) and the observation that changes in values of discrete-valued variables, such as propositional variables, apparently cannot avoid changing at instants. This view is consistent with the approach we take in the modelling of continuous durative actions, and with the view of change as consisting of both discrete and continuous aspects (Henzinger, 1996).

In the remainder of this section we compare the PDDL extensions that we propose with previous work in temporal reasoning by considering the three central issues identified above. Our objective is not to claim that our extensions improve on previous work, but instead to demonstrate that the implementation of solutions to these three problems within the PDDL framework makes their exploitation directly accessible to planning in a way that they are not when embedded within a logic and accompanying proof theory.

## 11.1 Continuous change

Several temporal reasoning frameworks began with consideration of discrete change and, later, were extended to handle continuous change. For example, Shanahan (1990) extended the event calculus of Kowalski and Sergot (1986) to enable the modelling of continuous change. This process of extension mirrors the situation faced in extending PDDL, where a system modelling discrete change already existed. It is, therefore, interesting to compare the use of PDDL2.1 with the use of systems such as the extended event calculus.

In his sink-filling example Shanahan (1990) discusses the issues of termination of events (self-termination and termination by other events), identification of the level of water in the sink during the filling process and the effect on the rate of change in the level of water in a sink when it is being filled from two sources simultaneously. The behaviour of the filling process and its effects on the state of the sink over time are modelled as axioms which would allow an inference engine to predict the state of the sink at points during the execution of the process.

PDDL2.1 allows the representation of the complex interactions that arise when a sink is filled from multiple independently controlled water sources by means of concurrent durative actions with continuous effects that encapsulate the initiation of the filling process, from a single water source, the change in the level of water in the sink and the termination of the process when the water source is turned off. This model is robust, since it easily accommodates multiple water sources, simply modifying the rate of flow appropriately by commutative updates. Since the actions have additive effects and the model provides the rate at which water enters the tank from a source, it is possible to compute the level of water in the sink at any point in the filling interval at which a concurrent action might consult the level. In contrast to Shanahan's extension to the event calculus, this approach does not require that the filling process be (at least from the point of view of the logical axiomatisation) terminated and restarted at a new rate when a water source is opened or closed, since the process simply remains active throughout. The change in rate of filling is then reflected in a piecewise-linear profile for the depth of water in the sink, just as it is in Shanahan's model.

It is not possible to model the multiple water sources situation if the filling process is completely encapsulated within a discretized durative action. In the discretized action the true level of water is not accessible during the filling process but only at its end or its start. Step-function behaviour only coarsely approximates true behaviour, so the consequence is that complex interactions cannot be properly modelled.

One of the important consequences of continuous behaviour is the triggering of events. In Shanahan's extensions this is achieved through the axiomatisation of causal relationships — events are not distinguished syntactically from actions, but only by the fact that their happening is axiomatically the consequence of certain conditions. In PDDL2.1 some events (such as the flooding of the sink if the filling continues after its capacity has been reached) can be modelled by using a combination of conditional effects and duration inequalities. However, not all events can be modelled in this way, since it is not always possible to predict when spontaneous events will occur. PDDL2.1 could be extended to allow the expression of causal axioms, but an alternative approach is to modify the language to enable the representation of events within the action-oriented tradition. This can be achieved by breaking up continuous durative actions into their instantaneous start and end points and the processes they encapsulate. This would enable the execution of a process to be initiated by a start action and ended by an instantaneous state transition that is either an action under the control of the planner or an *event*. A simple extension to the language is needed to distinguish actions from events and to prevent the planner from deliberately selecting an event. We refer to this approach as the *start-process-stop* model, and we have extended PDDL2.1 to support it (Fox & Long, 2002). The resulting language, PDDL+, is more difficult to plan with than PDDL2.1, and there are still open questions, concerning the complexity of the plan validation problem for this language, which remain topics for future work.

## 11.2 Concurrency

The opportunity for concurrent activities complicates several aspects of temporal reasoning. Firstly, it is necessary to account for which actions can be concurrent and secondly it is necessary to describe how concurrent activities interact in their effects on the world.

In most formalisms the first of these points is achieved by relying on the underlying logic to deliver an inconsistency when an attempt is made to apply two incompatible actions simultaneously. For example, the axioms of the event calculus will yield the simultaneous truth and falsity of a fluent if incompatible actions are applied simultaneously and consequently yield an inconsistency. Unfortunately, recognising inconsistency is, in general, undecidable, for a sufficiently expressive language. In PDDL2.1 we adopt a solution that exploits the restricted form of the action-centred formalism, defining the circumstances in which two actions could lead to inconsistency and rejecting the simultaneous application of such actions. We favour a conservative restriction on compatibility of actions (the *no moving targets* rule), in order to support efficient determination of incompatibility, rather than a more permissive but elusive ruling. An alternative approach, adopted by Bacchus in TLplan (2001), for example, is to allow multiple actions to occur at the same instant, but nevertheless to be executed in sequence. We find this solution counter-intuitive and, more importantly, consider that it would be impossible to use a plan of this sort as an instruction to an executive — no executive could be equipped to execute actions simultaneously and yet in a specified order. Our view is that if the order of execution matters then the executive must ensure that the actions are sequenced and can only do so within the limitations of its capability to measure time and react to its passing.

Shanahan (1999) discusses Gelfond's (1991) example of the soup bowl in which the problem concerns raising a soup bowl without spilling the soup. Two actions, lift left and lift right, can be applied to the bowl. If either is applied on its own the soup will spill, but, it is argued, if they are applied simultaneously then the bowl is raised from the table and no soup spills. Shanahan considers this example within the event calculus, where he uses an explicit assertion of the interaction between the lift left and lift right actions to ensure that the spillage effect is cancelled when the pair is executed together. The assumption is that the two actions can be executed at precisely the same moment and that the reasoner can rely on the successful simultaneity in order to exploit the effect.

In PDDL2.1 we take the view that precise simultaneity is outside the control of any physical executive. A plan is interpreted as an instruction to some executive system and we hold that no executive system is capable of measuring time and controlling its activity at arbitrarily fine degrees of accuracy. In particular, it is not possible for an executive to ensure that two actions that must be independently initiated are executed simultaneously. If a plan were to rely on such precision in measurement then, we claim, it could not be executed with any reliable expectation of success and should not, therefore, be considered a valid plan.

PDDL2.1 supports the modelling of the soup bowl situation in the following way. Two durative actions, *lift left* and *lift right*, both independently initiate tilting intervals which, when complete, will result in spillage of the soup if their effects have not been counteracted. Provided that the two lift actions start within an appropriate tolerance of one another the tilting will be corrected and the spillage avoided without the need to model cancellation of effects. We argue that an executive can execute the two actions to within a fine but non-zero tolerance of one another, and can therefore successfully lift the bowl. The event calculus model presented by Shanahan insists on precise synchronization of the two actions, incorrectly allowing it to be inferred that the soup will be spilled even if the time that elapses between the two lifts is actually small enough to allow for correction of the tilting of the bowl. Worse, Shanahan's axioms would allow lack of precise synchronization to be exploited to achieve spillage, using an amount of time smaller than that correctly describing the physical situation being modelled.

If one considers it unnecessary to model the precise interaction between the two lifts, one has the alternative in PDDL2.1 to abstract out the interaction and see the soup-bowl lifting action as a single discretized action that achieves the successful raising of the bowl.

## 11.3 Temporal extent

A common concern in temporal reasoning frameworks, discussed in detail by Vila and others (Vila, 1994; van Bentham, 1983), is the *divided instant problem*. This is the problem that is apparent when considering what happens at the moment of transition from, say, truth to falsity of a propositional variable. The question that must be addressed is whether the proposition is true, false, undefined or inconsistently both true and false at the instant of transition. Clearly the last of these possibilities is undesirable. The solution we adopt is a combination of the pragmatic and the philosophically principled. The pragmatic element is that we choose to model actions as instantaneous transitions with effects beginning at the instant of application. Thus, the actions mark the end-points of intervals of persistence of state which are closed on the left and open on the right. This ensures that the intervals nest together without inconsistency and the truth values of propositions are always defined. The same half-open-half-closed solution is adopted elsewhere. For example, Shanahan (1999) observes that a similar approach is used in the event calculus, although there the intervals are closed on the right. Although the two choices are effectively equivalent, we slightly prefer the closed-on-the-left choice since this allows the validation of a plan to conclude with the state at the point of execution of its final action, making the determination of the temporal span of the plan unambiguous.

From a philosophical point of view the truth value of the proposition at the instant of application of an action cannot be exploited by any other action, by virtue both of the no moving targets rule and our position, outlined above, that a valid plan cannot depend on precise synchronisation of actions. This forces actions that require a proposition as a precondition to sit at the open end of a half-open interval in which the proposition holds.

## 11.4 Planning with Time

In classical planning models, time is treated as relative. That is, the only temporal structuring in a plan, and in reasoning about a plan, is in the ordering between actions. This is most clearly emphasised by the issues that dominated planning research in the late 1980s and early 1990s, when classical planning was mainly characterised by the exploration of partial plan spaces, in planners such as TWEAK (Chapman, 1987), SNLP (McAllester & Rosenblitt, 1991) and UCPOP (Penberthy & Weld, 1992). Partial plans include a collection of actions representing the activity thus far determined to be part of a possible plan and a set of temporal constraints on those actions. The temporal constraints used in a partial plan are all of the form $A < B$ where $A$ and $B$ are time points corresponding to the application of actions.

Classical linear planners (Fikes & Nilsson, 1971; Russell & Norvig, 1995) rely on the simple fact that a total ordering on the points at which actions are applied can be trivially embedded into a time line. Again, the duration between actions is not considered. The role of time in planning becomes far more significant once metric time is introduced. With metric time it is possible to associate specific durations with actions, to set deadlines or windows of opportunity. The problems associated with relative time have still to be resolved in a metric time framework, but new problems are introduced. In particular, durations become explicit, so it is necessary to decide what the durations attach to: actions or states. Further, explicit temporal extents make it more important to confront the issue of concurrency in order to best exploit the measured temporal resources available to a planner.

In contrast to the simple ordering constraints required for relative time, metric time requires more powerful constraint management. Most metric time constraint handlers are built around the foundations laid by Dechter, Meiri and Pearl (1991). For example, IxTeT uses extensions of temporal constraint networks (Laborie & Ghallab, 1995). The language that IxTeT uses to represent planning domains is similar to PDDL2.1 as described in this paper, but more expressive because it allows access to time points within the interval of a durative action. This added expressive power is obtained at the cost of increased semantic complexity and, consequently, increased difficulty in the validation of plans. However, there are many similarities between the modelling of discretised durative actions in PDDL2.1 and in IxTeT, and similar modelling conventions are also found in the languages of Sapa (Do & Kambhampati, 2001) and Oplan (Drabble & Tate, 1994).

One of the earliest planners to consider the use of metric time was Deviser (Vere, 1983), which was developed from NONLIN (Tate, 1977). In Deviser, metric constraints on the times at which actions could be applied and deadlines for the achievements of goals were both expressible and the planner could construct plans respecting metric temporal constraints on the interactions between actions. Cesta and Oddi (1996) have explored various developments of temporal constraint network algorithms to achieve efficient implementation for planning and Galipienso and Sanchis (2002) consider extensions to manage disjunctive temporal constraints efficiently, which is a particularly valuable expressive element for plan construction as was observed above, since constraints preventing overlap of intervals translate into disjunctive constraints on time points. HSTS (Muscettola, 1994) also relies on a temporal constraint manager.

In systems that use continuous real-valued time it is possible to make use of linear constraint solvers to handle temporal constraints. In particular, constraints dictated by the relative placement of actions with durations on a timeline can be approached in this way (Long & Fox, 2003a). An alternative timeline that is often used is a discretised line based on integers. The advantage of this approach is that it is possible to advance time to a *next* value after considering activity at any given time point. The *next* modality can be interpreted in a continuous time framework by taking it to mean the state following the next logical change, regardless of the time at which this occurs (Bacchus & Kabanza, 1998). In planning problems in which no events can occur other than the actions dictated by the planner and no continuous change is modelled, plans are finite structures and therefore change can occur at only a finite number of time points during its execution. This makes it possible to embed the execution of the plan into the integer-valued discrete time line without any loss of expressiveness.

Various researchers have considered the problem of modelling continuous change. Pednault (1986) proposes explicit description of the functions that govern the continuous change of metric parameters, attached to actions that effect instantaneous change to initiate the processes. However, his approach is not easy to use in describing interacting continuous processes. For example, if water is filling a tank at a constant rate and then an additional water source is added to increase the rate of filling then the action initiating the second process must combine the effects of the two water sources. This means that the second action cannot be described simply in terms of its direct effect on the world — to increase the rate of flow into the tank — but with reference to the effects of other actions that have already affected the rate of change of the parameter. Shanahan (1990) also uses this approach, with the consequence that processes are modelled as stopping and then restarting with new trajectories as each interacting action is applied.

In Zeno (Penberthy & Weld, 1994), actions have effects that are described in terms of derivatives. This approach makes it easier to describe interacting processes, but complicates the management of processes by making it necessary to solve differential equations. The complication has not deterred other authors from taking this approach: McDermott (2003) takes this approach in his process planner.

The introduction of continuous processes into the planning problem represents a considerable complication, even over a model that includes temporal features and supports concurrency. It is an area of active research and the community has not yet agreed on matters of representation, let alone semantics. There remain many open problems for the planning community to address, both in the development of languages and planning algorithms and also in in the development of plan verification tools that can embody a widely accepted semantics.

## 12. Conclusions

Recent developments in AI planning research have been leading the community closer to the application of planning technology to realistic problems. This has necessitated the development of a representation language capable of modelling domains with temporal and metric features. The approach we have taken towards the development of such a language is to extend McDermott's PDDL domain representation standard to support temporal and metric models.

The development of the PDDL sequence towards greater expressive power is important to the planning community because PDDL has provided a common foundation for a great deal of recent research effort. The problems involved in modelling the behaviour of domains with both discrete and continuous behaviours have been well explored in the temporal logic and model checking communities but there have been no widely adopted models within the planning community. Our work on PDDL2.1 provides a way of making the relevant developments in these communities accessible to planning. Furthermore, PDDL2.1 begins to bridge the gap between basic research and applications-oriented planning by providing the expressive power necessary to capture real problems.

PDDL2.1 has the expressive power to represent a class of deterministic mixed discrete-continuous domains as planning domains. The language introduces a form of durative action based on three connected parts: the initiation of an interval in which numeric change might occur and its explicit termination by means of an action that produces the state corresponding to the end of the durative interval. This form of action allows the modelling of both discrete and continuous behaviours — discretized change can be represented by means of step functions, whilst continuous change can be modelled using the $\#t$ variable. The language provides solutions to the critical issues of concurrency, continuous change and temporal extent. The semantics of the language is derived from the familiar state transition semantics of STRIPS, extended to interpret invariants holding over intervals in which continuous functions might also be active. Our semantics allows us to interpret more plans than we can efficiently validate. We describe the criteria that a plan must satisfy in order to be practically verifiable.

This paper has focussed primarily on a discussion of the numeric and discretised temporal features of PDDL2.1. However, the modelling capability of discretized durative actions is in some respects limited and it is important for the planning community to address the challenges presented by continuous change. Indeed, even using the continuous actions of PDDL2.1 it is not possible to model episodes of change being terminated by spontaneous events in the world rather than by the deliberate choice of the planner. The future goals of the community should include addressing domains that require the continuous actions of PDDL2.1, then confronting the challenges of planning within more dynamic environments in which intervals of change can be terminated by the world as well as by the deliberate action of the planner. This will constitute an important step towards planning within dynamic and unpredictable environments.

## Acknowledgements

We would like to thank the members of the committee for the third International Planning Competition. In particular, discussions with Drew McDermott, Fahiem Bacchus, David Smith and Hector Geffner in turns infuriated, intrigued and delighted us and contributed immeasurably to the strengths of this paper. Many others have offered comments and insights that have allowed us to develop the work we present here. We would like to thank Jörg Hoffmann, Malte Helmert, Antonio Garrido, Stefan Edelkamp, Nicola Muscettola, Mark Boddy, Keith Golden, Jeremy Frank, Ari Jónsson, Julie Porteous, Alex Coddington, Stephen Cresswell, Luke Murray, Keith Halsey and Richard Howey for the many helpful discussions we have shared.

## Appendix A. BNF Specification of PDDL2.1

This appendix contains a complete BNF specification of the PDDL2.1 language. This is not
a strict superset of PDDL1.x. For example, the use of local variables within action schemas
has been left out of this specification. It is not a widely used part of the language and has
not been used in any of the competition domains. The interpretation of local variables as
proposed by McDermott is subtle, since it demands confirmation that a unique instantiation
exists for each such variable. It is non-trivial to confirm that this is the case during plan
validation for domains with significant expressive power and the fact that it has been largely
ignored suggests that it is poorly understood. Other changes are discussed in the following
sections.

### A.1 Domains

Domain structures remain essentially as specified in PDDL1.x. The main alterations are
to introduce a slightly modified syntax for numeric fluent expressions and to remove the
syntax for hierarchical expansions. The latter is not necessarily abandoned, but it has not,
to the best of our knowledge, been used in any publicly available planning systems or even
domains. In the original PDDL specification, a distinction was drawn between *strict* PDDL
and *non-strict* PDDL, where strict PDDL must follow the ordering of the fields specified below,
while non-strict PDDL is not restricted in this way. In practice, there are relatively few fields
that it is intuitive to accept in arbitrary orders — it is natural to expect declarations to
precede use of symbols and for preconditions to precede effects. However, declarations of
constants, predicates and function symbols are not naturally ordered, so in the current
definition of PDDL the ordering of all fields must follow the specification below, with the
exception of these three fields which are legal in any order with respect to one another,
although the group must follow types (if there are any) and precede action specifications.

```
<domain>              ::= (define (domain <name>)
                            [<require-def>]
                            [<types-def>]:typing
                            [<constants-def>]
                            [<predicates-def>]
                            [<functions-def>]:fluents
                            <structure-def>*)
<require-def>         ::= (:requirements <require-key>+)
<require-key>         ::= See Section A.5
<types-def>           ::= (:types <typed list (name)>)
<constants-def>       ::= (:constants <typed list (name)>)
<predicates-def>      ::= (:predicates <atomic formula skeleton>+)
<atomic formula skeleton>
                      ::= (<predicate> <typed list (variable)>)
<predicate>           ::= <name>
<variable>            ::= ?<name>
<atomic function skeleton>
                      ::= (<function-symbol> <typed list (variable)>)
<function-symbol>     ::= <name>
<functions-def>       ::=:fluents (:functions <function typed list
                              (atomic function skeleton)>)
<structure-def>       ::= <action-def>
<structure-def>       ::=:durative-actions <durative-action-def>
```

A slight modification has been made to the type syntax – it is no longer possible to nest `either` expressions (a possibility that was never exploited, but complicates parsing). Numbers are no longer considered to be an implicit type – the extension to numbers is now handled only through functional expressions. This ensures that there are only finitely many ground action instances. A desirable consequence is that action selection choice points need never include choice over arbitrary numeric ranges. The use of finite ranges of integers for specifying actions is useful (see Mystery or FreeCell for example) and an extension of the standard syntax to allow for a more convenient representation of these cases could be useful. The syntax of function declarations allows functions to be declared with types. At present the syntax is restricted to number types, since we do not have a semantics for other functions, but the syntax offers scope for possible extension. Where types are not given for the function results they are assumed to be numbers.

```
<typed list (x)> ::= x*
<typed list (x)> ::=:typing x+- <type> <typed list(x)>
<primitive-type> ::= <name>
<type>           ::= (either <primitive-type>+)
<type>           ::= <primitive-type>

<function typed list (x)> ::= x*
<function typed list (x)> ::=:typing x+- <function type>
                                       <function typed list(x)>
<function type>            ::= number
```

## A.2 Actions

The BNF for an action definition is given below. Again, this has been simplified by removing generally unused constructs (mainly hierarchical expansions). It should be emphasised that this removal is not intended to be a permanent exclusion — hierarchical expansion syntax has proved a difficult element of the language both to agree on and to exploit. As the other levels of the language stabilise we hope to return to this layer and redevelop it.

```
<action-def>      ::= (:action <action-symbol>
                          :parameters ( <typed list (variable)> )
                          <action-def body>)
<action-symbol>   ::= <name>
<action-def body> ::= [:precondition <GD>]
                      [:effect <effect>]
```

Goal descriptions have been extended to include fluent expressions.

```
<GD>              ::= ()
<GD>              ::= <atomic formula(term)>
<GD>              ::=:negative-preconditions <literal(term)>
<GD>              ::= (and <GD>*)
```

```
<GD>                  ::=:disjunctive−preconditions (or <GD>*)
<GD>                  ::=:disjunctive−preconditions (not <GD>)
<GD>                  ::=:disjunctive−preconditions (imply <GD> <GD>)
<GD>                  ::=:existential−preconditions
                         (exists (<typed list(variable)>*) <GD> )
<GD>                  ::=:universal−preconditions
                         (forall (<typed list(variable)>*) <GD> )
<GD>                  ::=:fluents <f-comp>
<f-comp>              ::= (<binary-comp> <f-exp> <f-exp>)
<literal(t)>          ::= <atomic formula(t)>
<literal(t)>          ::= (not <atomic formula(t)>)
<atomic formula(t)>   ::= (<predicate> t*)
<term>                ::= <name>
<term>                ::= <variable>
<f-exp>               ::= <number>
<f-exp>               ::= (<binary-op> <f-exp> <f-exp>)
<f-exp>               ::= (- <f-exp>)
<f-exp>               ::= <f-head>
<f-head>              ::= (<function-symbol> <term>*)
<f-head>              ::= <function-symbol>
<binary-op>           ::= +
<binary-op>           ::= −
<binary-op>           ::= *
<binary-op>           ::= /
<binary-comp>         ::= >
<binary-comp>         ::= <
<binary-comp>         ::= =
<binary-comp>         ::= >=
<binary-comp>         ::= <=
<number>              ::= Any numeric literal
                         (integers and floats of form n.n).
```

Effects have been extended to include functional expression updates. The syntax proposed here is a little different from the syntax proposed in the earlier version of PDDL. The syntax of conditional effects proposed by Fahiem Bacchus for AIPS 2000 has been adopted, in which the nesting of conditional effects is not supported. The assignment operators are prefix forms. Simple assignment is called `assign` (previously this was `change`) and operators corresponding to C update assignments, $+=, -=, *=$ and $/=$ are given the names `increase`, `decrease`, `scale-up` and `scale-down` respectively. The prefix form has been adopted in preference to an infix form in order to preserve consistency with the Lisp-like syntax and the non-C names to help the C and C++ programmers to remember that the operators are to be used in prefix form). We prefer `assign` to the original `change` because the introduction of `increase` and so on makes the nature of a *change* more ambiguous.

```
<effect>     ::= ()
<effect>     ::= (and <c-effect>*)
<effect>     ::= <c-effect>
<c-effect>   ::=:conditional−effects (forall (<variable>*) <effect>)
<c-effect>   ::=:conditional−effects (when <GD> <cond-effect>)
<c-effect>   ::= <p-effect>
<p-effect>   ::= (<assign-op> <f-head> <f-exp>)
<p-effect>   ::= (not <atomic formula(term)>)
<p-effect>   ::= <atomic formula(term)>
```

```
<p-effect>       ::=ᶠˡᵘᵉⁿᵗˢ(<assign-op> <f-head> <f-exp>)
<cond-effect> ::= (and <p-effect>*)
<cond-effect> ::= <p-effect>
<assign-op>   ::= assign
<assign-op>   ::= scale-up
<assign-op>   ::= scale-down
<assign-op>   ::= increase
<assign-op>   ::= decrease
```

## A.3 Durative Actions

Durative action syntax is built on a relatively conservative extension of the existing action syntax.

```
<durative-action-def> ::= (:durative-action <da-symbol>
                              :parameters ( <typed list (variable)> )
                              <da-def body>)
<da-symbol>           ::= <name>
<da-def body>         ::= :duration <duration-constraint>
                            :condition <da-GD>
                            :effect <da-effect>
```

The conditions under which a durative action can be executed are more complex than for standard actions, in that they specify more than the conditions that must hold at the point of execution. They also specify the conditions that must hold throughout the duration of the durative action and also at its termination. To distinguish these components we introduce a simple temporal qualifier for the preconditions. The use of the name "precondition" would be somewhat misleading given that the conditions described can include constraints on what must hold after the action has begun. This has motivated the adoption of :condition to describe the collection of constraints that must hold in order to successfully apply a durative action. The logical form of conditions for durative actions has been restricted to conjunctions of temporally annotated expressions, but there is clearly scope for future extension to allow more complex formulae.

```
<da-GD>             ::= ()
<da-GD>             ::= <timed-GD>
<da-GD>             ::= (and <timed-GD>⁺)
<timed-GD>          ::= (at <time-specifier> <GD>)
<timed-GD>          ::= (over <interval> <GD>)
<time-specifier>    ::= start
<time-specifier>    ::= end
<interval>          ::= all
```

The duration (?duration) of a durative action can be specified to be equal to a given expression (which can be a function of numeric expressions), or else it can be constrained with inequalities. This latter allows for actions where the conclusion of the action can be freely determined by the executive without necessarily having further side-effects. For example, a walk between two locations could be made to take as long as the executive considered convenient, provided it was at least as long as the time taken to walk between the locations at the fastest walking speed possible. Constraints that do not specify the exact duration of a durative action might prove harder to handle, so we have introduced a label (`:duration-inequalities`) to signal that a domain makes use of them. A duration constraint is supplied to dictate or limit the temporal extent of the durative action. The duration is an implicit parameter of the durative action and must be supplied in a plan that uses durative actions. To denote this, a durative action is denoted in a plan by `t:(name arg1...argn)[d]` where d is the (non-negative, rational valued) duration in floating point format ($n.n$). Duration constraints can be explicitly temporally annotated to indicate that they should be evaluated in the context of the start or end point of the action, or else they can be left unannotated, in which case the default is that they are evaluated in the context at the start of the action (as indicated in Definition 16).

```
<duration-constraint>              ::=:duration-inequalities
                                       (and <simple-duration-constraint>+)
<duration-constraint>              ::= ()
<duration-constraint>              ::= <simple-duration-constraint>
<simple-duration-constraint>       ::= (<d-op> ?duration <d-value>)
<simple-duration-constraint>       ::= (at <time-specifier>
                                          <simple-duration-constraint>)
<d-op>                             ::=:duration-inequalities <=
<d-op>                             ::=:duration-inequalities >=
<d-op>                             ::= =
<d-value>                         ::= <number>
<d-value>                         ::=:fluents <f-exp>
```

In addition to logical effects, which can occur at the start or end of a durative action, durative actions can have numeric effects that refer to the literal `?duration`. More sophisticated durative actions can also make use of functional expressions describing effects that occur over the duration of the action. This allows functional expressions to be updated by a continuous function of time, rather than only step functions.

```
<da-effect>     ::= ()
<da-effect>     ::= (and <da-effect>*)
<da-effect>     ::= <timed-effect>
<da-effect>     ::=:conditional-effects (forall (<variable>*) <da-effect>)
<da-effect>     ::=:conditional-effects (when <da-GD> <timed-effect>)
<da-effect>     ::=:fluents (<assign-op> <f-head> <f-exp-da>)
<timed-effect>  ::= (at <time-specifier> <a-effect>)
<timed-effect>  ::= (at <time-specifier> <f-assign-da>)
<timed-effect>  ::=:continuous-effects (<assign-op-t> <f-head> <f-exp-t>)
<f-assign-da>   ::= (<assign-op> <f-head> <f-exp-da>)
<f-exp-da>      ::= (<binary-op> <f-exp-da> <f-exp-da>)
<f-exp-da>      ::= (- <f-exp-da>)
<f-exp-da>      ::=:duration-inequalities ?duration
<f-exp-da>      ::= <f-exp>
```

Note that the `?duration` term can only be used to define functional expression updating effects if the duration constraints requirement is set. This is because in other cases the duration value is available as an expression, whereas when duration constraints are provided the duration can, sometimes, be freely selected within constrained boundaries.

```
<assign-op-t>        ::= increase
<assign-op-t>        ::= decrease
<f-exp-t>            ::= (* <f-exp> #t)
<f-exp-t>            ::= (* #t <f-exp>)
<f-exp-t>            ::= #t
```

The symbol #$t$ is used to represent the period of time that a given durative action has been active. It is therefore a *local* clock value for each duration, independent of similar clocks for each other duration. There has been discussion with members of the committee about the use of the expression using #$t$: it was proposed that an expression declaring the rate of change alone could be used. We decided against this on the grounds that the assertion of a rate of change suggests that the rate of change is determined by one process effect alone. In fact, it is intended that if multiple active processes affect the same fluent then these effects are accumulated. Using the expression that directly defines the amount by which each process contributes to the change in a fluent value over time we do not appear to assert (inconsistently) that a fluent has multiple simultaneous rates of change.

## A.4 Problems

Planning problems specifications have been modified to exclude several generally unused constructs (named initial situations and expansion information). We have removed the length specification because it is at odds with the intention to supply physics, not advice. Furthermore, the advice this field offers over-emphasises a very coarse plan metric. Instead, we have introduced an optional metric field, which can be used to supply an expression that should be optimized in the construction of a plan. The field states whether the metric is to be minimized or maximized. Of course, a planner is free to ignore this field and make the assumption that plans with fewest steps will be considered good plans. However, we consider this extension to be a crucial one in the development of a more widely applicable planning language. We have provided the variable `total-time` that takes the value of the total execution time for the plan. This allows us to conveniently express the intention to minimize total execution time.

We anticipate that extensions of the plan metric syntax will prove necessary in the longer term, but believe that this version already provides a significant new challenge to the community. Problem specifications are still somewhat impoverished in terms of the ability to easily specify temporal constraints on goals and other non-standard features of initial and goal states. Again, we anticipate the need for extension, but have chosen to leave a clean sheet for future developments.

```
<problem>            ::= (define (problem <name>)
                         (:domain <name>)
                         [<require-def>]
                         [<object declaration> ]
                         <init>
                         <goal>
                         [<metric-spec>]
                         [<length-spec> ])
<object declaration> ::= (:objects <typed list (name)>)
<init>               ::= (:init <init-el>*)
<init-el>            ::= <literal(name)>
```

```
<init-el>            ::=:fluents (= <f-head> <number>)
<goal>               ::= (:goal <GD>)
<metric-spec>        ::= (:metric <optimization> <ground-f-exp>)
<optimization>       ::= minimize
<optimization>       ::= maximize
<ground-f-exp>       ::= (<binary-op> <ground-f-exp> <ground-f-exp>)
<ground-f-exp>       ::= (- <ground-f-exp>)
<ground-f-exp>       ::= <number>
<ground-f-exp>       ::= (<function-symbol> <name>*)
<ground-f-exp>       ::= total-time
<ground-f-exp>       ::= <function-symbol>
<length-spec>        ::= (:length [(:serial <integer>)]
                            [(:parallel <integer>)])
```
*The length-spec is deprecated.*

## A.5 Requirements

Here is a table of all requirements in PDDL2.1. Some requirements imply others; some are abbreviations for common sets of requirements. If a domain stipulates no requirements, it is assumed to declare a requirement for `:strips`.

| Requirement | Description |
|---|---|
| :strips | Basic STRIPS-style adds and deletes |
| :typing | Allow type names in declarations of variables |
| :negative-preconditions | Allow not in goal descriptions |
| :disjunctive-preconditions | Allow or in goal descriptions |
| :equality | Support = as built-in predicate |
| :existential-preconditions | Allow exists in goal descriptions |
| :universal-preconditions | Allow forall in goal descriptions |
| :quantified-preconditions | = :existential-preconditions |
|  | + :universal-preconditions |
| :conditional-effects | Allow when in action effects |
| :fluents | Allow function definitions and use of effects using |
|  | assignment operators and arithmetic preconditions. |
| :adl | = :strips + :typing |
|  | + :negative-preconditions |
|  | + :disjunctive-preconditions |
|  | + :equality |
|  | + :quantified-preconditions |
|  | + :conditional-effects |
| :durative-actions | Allows durative actions. |
|  | Note that this does not imply :fluents. |
| :duration-inequalities | Allows duration constraints in durative |
|  | actions using inequalities. |
| :continuous-effects | Allows durative actions to affect fluents |
|  | continuously over the duration of the actions. |